\pdfoutput=1
\documentclass[authoryear,round, colorlinks=true, citecolor=blue, linkcolor=blue, urlcolor=blue,11pt]{article}

\usepackage[final]{acl}

\usepackage{times}
\usepackage{latexsym}
\usepackage{amsmath}
\usepackage[utf8]{inputenc} % allow utf-8 input
\usepackage[T1]{fontenc}    % use 8-bit T1 fonts
\usepackage{hyperref}       % hyperlinks
\usepackage{url}            % simple URL typesetting
\usepackage{booktabs}       % professional-quality tables
\usepackage{amsfonts}       % blackboard math symbols
\usepackage{nicefrac}       % compact symbols for 1/2, etc.
\usepackage{microtype}      % microtypography
\usepackage{xcolor}         % colors
\usepackage{amssymb}
\usepackage{comment}
\usepackage{multirow}
\usepackage{array}
\usepackage{graphicx}
\usepackage{subcaption}

\usepackage[T1]{fontenc}
\usepackage[utf8]{inputenc}

\usepackage{microtype}

\usepackage{inconsolata}

\usepackage{graphicx}

\title{A Unified Framework for Model Editing}

\author{Akshat Gupta, Dev Sajnani, Gopala Anumanchipalli \\
        UC Berkeley \\ 
        \texttt{\{akshat.gupta, sajnanidev, gopala\}@berkeley.edu}
        }

\begin{document}
\maketitle
\begin{abstract}
% ROME and MEMIT stand out as leading "locate-and-edit" model editing techniques. The optimization objectives of the two algorithms are different - ROME uses an equality constraint to perform one edit at a time, whereas MEMIT employs a more flexible least-square constraint, thus allowing for batched edits. 
ROME and MEMIT are largely believed to be two different model editing algorithms, with the major difference between them being the ability to perform batched edits. In this paper, we unify these two algorithms under a single conceptual umbrella, optimizing for the same goal, which we call the \textbf{preservation-memorization} objective. ROME uses an equality constraint to optimize this objective to perform one edit at a time, whereas MEMIT employs a more flexible least-square constraint that allows for batched edits. We generalize ROME and enable batched editing with equality constraint in the form of \textbf{EMMET} - an \textbf{\underline{E}}quality-constrained \textbf{\underline{M}}ass \textbf{\underline{M}}odel \textbf{\underline{E}}diting algorithm for \textbf{\underline{T}}ransformers, a new batched memory-editing algorithm. EMMET can perform batched-edits up to a batch-size of 10,000, with very similar performance to MEMIT across multiple dimensions. With the introduction of EMMET, we truly unify ROME and MEMIT and show that both algorithms are equivalent in terms of their optimization objective, their abilities (singular and batched editing), their model editing performance and their limitations.

\end{abstract}

\section{Introduction}

As new facts emerge constantly, it is crucial to keep models up-to-date with the latest knowledge. Model editing \cite{editing-survey} gives us the ability to edit facts stored inside a model as well as update incorrectly stored facts. In this paper, we focus on two of the most popular and best performing model editing methods - ROME (Rank-One Model Editing) \citep{ROME} and MEMIT (Mass Editing Memory in Transformer) \citep{MEMIT}. ROME and MEMIT directly update specific "knowledge-containing" parts of the model without requiring the need to train additional models \citep{metamodel, MEND, MALMEN} and can be applied to any transformer based large language model (LLMs). This makes these algorithms really attractive for practical use cases. MEMIT also uniquely allows for \textit{batched edits} (appendix \ref{sec:related work}). %We refer the reader to section \ref{sec:related work} for a more detailed introduction to model editing.

% Model editing methods can be broadly classified into two types - methods that add information in-context \citep{SERAC, mquake, ripple-effects}, and methods that modify the parameters of underlying model \citep{metamodel, MEND, ROME, MEMIT, MALMEN}. Various model editing techniques have been proposed in the past that tackle this problem in different ways. \cite{knowledgeneurons} first identify knowledge containing neurons in a model using integrated gradients \citep{integrated-gradients} and then modify the selected neurons to edit facts in a model. This method is not scalable with increasing model sizes as it requires us to find activations for each neuron in the model. \cite{metamodel} and \cite{MEND} train a hypernetwork \citep{hypernetwork} that generates the new weights of the model being edited. While these methods have been optimized to scale with a square-root dependence on the size of the edited model, it still requires training of additional editing models dependent on each source model being edited. Other methods add the most relevant updated knowledge in context \citep{SERAC, ripple-effects, mquake}. While such methods provide a viable alternative to model editing, in this paper, we focus on parameter-modifying model editing methods.

\begin{figure*}
    \centering
    \includegraphics[width=0.8\linewidth]{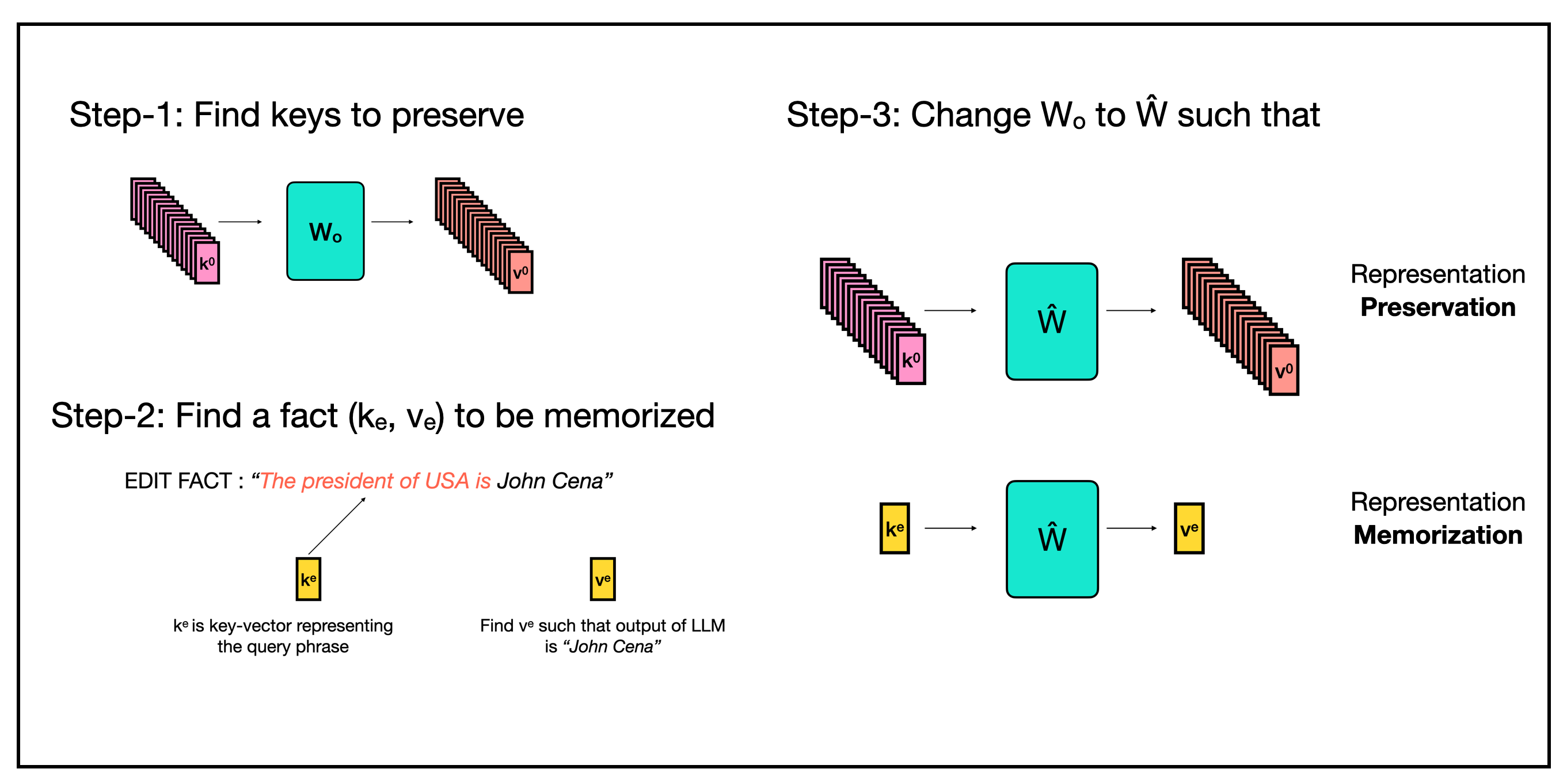}
    \caption{A diagrammatic representation of the preservation-memorization objective. }
    \label{fig:pm-objective}
\end{figure*}

%Write the summary of contributions here
ROME and MEMIT are largely considered different from each other, with one of their major differences being that ROME allows for editing only one fact at a time. In this paper, we present a unifying conceptual framework for ROME and MEMIT and show that both methods optimize the same objective function. We call this the \textbf{preservation-memorization} objective of model editing, where new knowledge is injected or memorized such that representations of certain vectors are preserved through the editing process. We show that ROME optimizes an equality-constrained version of the objective whereas MEMIT optimizes a more relaxed least-squares version of the objective, which allows for a simple closed-form solution for making batched edits. We then highlight that MEMIT consists of two separate steps - an optimization objective and an algorithm that distributes the edits into multiple layers. The power of MEMIT in many cases comes from these \textbf{edit-distribution} algorithms. %However, our experiments show that while MEMIT's edit-distribution algorithm helps editing with large batches for GPT2-XL \citep{gpt-2} and GPT-J \citep{gpt-j}, it hurts model editing performance when used with Llama-2-7b \citep{llama2}. With this, we advocate for further research into the edit-distribution algorithms and viewing them as separate entities from the optimization objectives. %RE-WRITE LAST PART OF WHAT YOU WANT TO DO WITH THIS AFTER RESULTS COME OUT

Finally, we present a closed-form solution for making batched edits with equality-constraint under the preservation-memorization objective in the form of EMMET - an \textbf{\underline{E}}quality-constrained \textbf{\underline{M}}ass \textbf{\underline{M}}odel \textbf{\underline{E}}diting algorithm for \textbf{\underline{T}}ransformers. With EMMET, batched edits can be performed for batch sizes up to 10,000 with performance much similar to MEMIT. We evaluate EMMET on three models - GPT2-XL, GPT-J and Llama-2-7b on standard model editing datasets - CounterFact and zsRE. Enabling batched editing with equality-constraint in the form of EMMET allows us to truly unify the two algorithms and shows that both ROME and MEMIT are essentially equivalent in terms of their optimization objective, their abilities (performing singular and batched editing), their model editing performance and their limitations. EMMET serves as a cornerstone in completing this larger picture. The code for EMMET can be found {here}\footnote{\url{https://github.com/scalable-model-editing/unified-model-editing}}.

The main contributions of our paper are:
\begin{itemize}
    \item We unify two popular model editing techniques (ROME and MEMIT) under the preservation-memorization objective and show that these algorithms are equivalent in terms of their optimization objective and in practice.
    \item We disentangle the MEMIT objective from the MEMIT algorithm which distributes edits within multiple layers. This allows for a fair comparison of MEMIT and ROME.
    \item We present a closed-form solution to equality-constrained memorization in the form of EMMET, a batched version of ROME. EMMET is a new batched-editing algorithm that achieves symmetry in usage and performance between the two algorithms and shows that batched edits can be made using both objectives. 

\end{itemize}
%While EMMET does batched editing under an equality constraint, enforcing a "theoretically" more accurate editing of memories, in practice, we find that EMMET performs on par with MEMIT. We believe the reason for this is the hard equality constraints enforced in the optimization objective of EMMET. While EMMET performance does not match MEMIT for larger batch sizes, it can still be used for sequential model editing with smaller batch sizes \citep{editing-survey, akshat-catastrophic}. 

\section{Background}

Facts for model editing are usually represented in a key-value format where the key vector has maximal correspondence to retrieval of a fact and the value vector enables us to get the target output after editing \citep{ROME, key-value-memories}. As an example, let us say we are editing a new fact into the model - \textit{"The president of USA is John Cena"}. In this fact, $k_e$ is the vector representation of the phrase - "The president of USA is," and $v_e$ is the vector representation of the output at the layer being edited such that "John Cena" is produced as output at the final layer of the model. This is pictorially represented in step 2 in Figure \ref{fig:pm-objective}. For a more detailed explanation of the creation of key-value vectors, we refer readers to \citep{ROME}.

%While early model editing methods made singular but accurate edits, more recently, focus has shifted to scaling up model editing methods. Scaling of model editing can happen in two ways - either through sequential editing \citep{editing-survey, akshat-catastrophic} or batched editing \citep{MEMIT}. In sequential editing, edits are made to the model sequentially whereas in batched editing, multiple facts are inserted into the model with a single gradient update. MEMIT \citep{MEMIT} is one such batched editing algorithm. In this paper, we present EMMET, which is a new batched editing algorithm that inserts multiple facts into the model with one gradient update. 

The success of model editing is measured using standard model editing metrics \citep{ROME, editing-survey} described below: 

\begin{figure*}
    \centering
        \includegraphics[width=0.9\linewidth]{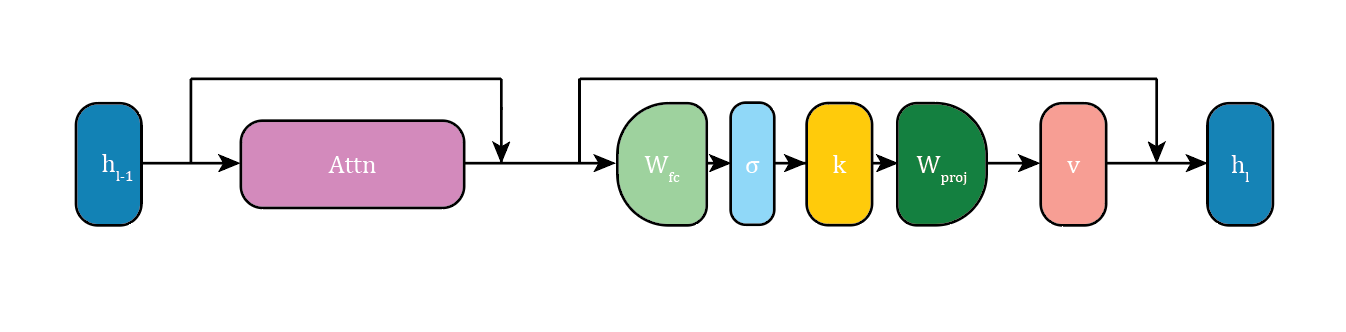}
    \caption{Figure shows a diagrammatic representation of a transformer layer. The layer being edited by ROME, MEMIT and EMMET is the projection weight matrix inside the MLP layer ($W_{proj}$).}
    \label{fig:diagram}
\end{figure*}

\begin{itemize}
    \item \textbf{Efficacy Score (ES)} indicates if an edit has been successfully made to a model. It is measured as the percentage of edits where $P(\text{new fact}) > P(\text{old fact})$ for a query prompt used to edit the model.
    
    \item \textbf{Paraphrase Score (PS)} represents the generalization ability of model under an edit. It is measured as the percentage of edits where $P(\text{new fact}) > P(\text{old fact})$ under paraphrases of the query prompt.

    \item \textbf{Neighborhood Score (NS)} represents locality of model editing. In other words, it measures if editing of a fact affects other facts stored inside a model. NS represents the percentage of facts in the neighborhood of the edited fact that remain unaltered post-edit.

    \item \textbf{Generation Entropy (GE)} represents the fluency of a model post edit. It is calculated by measuring the weighted average of bi-gram and tri-gram entropies of text generated by an edited model. This quantity drops if the generated text is repetitive, a common failure case of model editing \citep{ROME, akshat-rebuilding}.

    \item \textbf{Score (S)} is a quantity defined by \citep{ROME} to represent a combination of edit success, generalization and locality. It is the harmonic mean of ES, PS, and NS.
\end{itemize}

% \begin{comment}
% 1. Defining model editing and give some examples.
% 2. Define batch editing
% 3. Defining and describe different metrics used when evaluating model editing. (the metrics defined in ROME paper)
% 4. Describe datasets used in model editing
% 5. Facts as key-value pairs
% \end{comment}

\section{Preservation-Memorization : A Unifying Framework for ROME and MEMIT}\label{sec:pm}

% ROME and MEMIT are two of the most popular model editing methods. While ROME allows only singular edits, MEMIT allows for editing multiple memories in a model at the same time, thus providing a scalable solution to model editing. 

Both ROME and MEMIT base their work on viewing the weights of the feed-forward layer in a transformer as linear associative memories \citep{linearmemory1, linearmemory2}. Under this paradigm, linear operations in a transformer (feed-forward layers) are viewed as a key-value store for information. In this section, we re-introduce both ROME and MEMIT in a new light - a unifying conceptual framework of the \textbf{preservation-memorization} objective. %Specifically, the projection layer inside a transformer model are edited by these methods, as shown in Figure \ref{fig:diagram-full-layer}.

%The output of a single transformer layer for a general decoder-only LLM can be written as:

\begin{comment}
    
\begin{align}
    a^l &= {LN}(\texttt{Att}(h^{l-1}) + h^{l-1})\\
    m^l &= W^l_{proj} \left ( {NL}(W^l_{fc}a^{l}  + b^l_{fc}) \right )   + b^l_{proj}\\
    h^l &= {LN}(m^l + a^l)
\end{align}
\end{comment}

%Here, \texttt{Att} is the multi-head attention function, \texttt{NL} refers to the non-linearity used in the MLP module of the model and \texttt{LN} refers to layer normalization. The keys are outputs of the first linear layer, or $k^l = \texttt{NL}(a^{l} W^l_{fc}  + b^l_{fc})$, whereas $v^l = m^l$. A detailed explanation on creation of key-vectors and value-vectors is given in Appendix \ref{sec:key-value-creation}.

Let $W$ represent the weights of the feed-forward layer we want to edit\footnote{These layers are found by causal tracing methods \citep{ROME, MEMIT}}, and let $k$ be a key-vector representative of a fact that we are either editing or preserving, and is the input vector to $W$. The layers being edited are shown in an expanded diagram of a transformer layer \cite{transformers} in Figure \ref{fig:diagram}. In the model editing process, the weights of an intermediate layer of the model are changed from $W_0$ to $\hat{W}$ ($W_0$ represents the original weights of the $W_{proj}$ matrix), where $k_0$ is used to indicate a key-vector representing facts we want to preserve from the original model, and $k_e$ being key-vectors representing facts we want to insert into the model. Let $v_e$ be the desired output at the layer being edited corresponding to input $k_e$ such that the correct fact is recalled by the model when finally generating text. A detailed explanation on creation of key-vectors and value-vectors is given in Appendix \ref{sec:key-value-creation} and is also briefly depicted in Figure \ref{fig:pm-objective}.  %For more details on how we find $k_e$ and $v_e$, we refer the reader to \cite{ROME}.

Our objective is then to preserve the representations of selected input vectors before and after editing, or in other words, minimize the error between $W_0k_0$ and $\hat{W}k_0$, while forcing the output representation of the vector $k_e$ to be $v_e$, or in other words - memorizing the fact represented by ($k_e$, $v_e$). This process is shown pictorially in Figure \ref{fig:pm-objective}. 

In ROME-style, this objective of model editing is optimized by the following equation:

\begin{equation}
\begin{aligned}
    & \underset{\hat{W}}{\operatorname{argmin}} \hspace{4pt} \underbrace{\left\| \hat{W} K_0 - W_0 K_0 \right\|^2_F}_{\text{preservation}} \hspace{4pt}  \text{    s.t. } \underbrace{\hat{W} k_e = v_e}_{\text{memorization}}
\end{aligned}
\end{equation}

where $K_0 = [k^0_1 \hspace{4pt}| k^0_2 \hspace{4pt}| \dots |\hspace{4pt}k^0_N]$ is a matrix containing all the vectors whose representations we want to preserve in a row.

We call this the {preservation-memorization} objective of model editing because it allows us to retain existing knowledge or skills of a model {by keeping the same representations of selected key-vectors before and after editing, while memorizing a new fact $k_e$, whose representation are forced to be $v_e$,} where $v_e$ is by definition the output representation for $k_e$ that generates the target answer at final layer. 

The solution for ROME can then be written as:

\begin{align}\label{eq:rome-update-equation}
    \hat{W} &= W_0 + \Delta \hspace{10pt} \text{where} \hspace{10pt}\\
    \Delta &= (v_e - W_0k_e) \frac{k_e^TC_0^{-1}}{k_e^TC_0^{-1}k_e}
\end{align}

Here, $C_0 = K_0K_0^T$ is assumed to be an invertible matrix and the denominator $k_e^TC_0^{-1}k_e$ is a scalar. 

MEMIT on the other hand optimizes a relaxed version of the same objective:

\begin{equation}\label{eq:memit_objective}
     \underset{\hat{W}}{\operatorname{argmin}} \hspace{4pt} \underbrace{\lambda \left\| \hat{W} K_0 - W_0 K_0 \right\|^2_F}_{\text{preservation}}  + \underbrace{\left\|\hat{W} K_E - V_E \right\|^2_F}_{\text{memorization}}
\end{equation}

where $K_E = [k^e_1 \hspace{4pt}| k^e_2 \hspace{4pt}| \dots |\hspace{4pt}k^e_E]$ is a matrix containing a row of vectors representing the edits we are making in a batch and $V_E = [v^e_1 \hspace{4pt}| v^e_2 \hspace{4pt}| \dots |\hspace{4pt}v^e_E]$ represents their target representations. 
%We again see that the first term in the above equation preserves the representation of selected vectors while the second term forces memorization of facts represented by $(K_E, V_E)$ using a least square constraint.

\begin{table*}
    \vskip 0.05in
    \centering 
    \scriptsize
    % \small
    % {\textwidth}{@{\extracolsep{\fill}}*{11}{c}}
    \setlength\tabcolsep{0pt}
    \setlength\extrarowheight{1pt}
    % \begin{tabular}{c c c c c c c c c c c}
    \begin{tabular*}{\textwidth}{@{\extracolsep{\fill\centering}}*{10}{c}}
        \toprule 
        % \multicolumn{12}{c}{DATASET: MPI-1k $\vert$ MODEL: \textsc{OPT}}\\
        % \midrule
        \multirow{2}{*}[-0.3em]{\textsc{Algorithm}} & 
            \multirow{2}{*}[-0.3em]{\textsc{Model}} &
            \multicolumn{2}{c}{Efficacy} & 
            \multicolumn{2}{c}{Generalization} & 
            \multicolumn{2}{c}{Locality} & 
            \multicolumn{1}{c}{Fluency} & 
            \multicolumn{1}{c}{Score}\\ 
        \addlinespace[0.125em] \cline{3-10} \addlinespace[0.25em]
        &  & ES $\uparrow$ & EM $\uparrow$ &
             PS $\uparrow$ & PM $\uparrow$ &
             NS $\uparrow$ & NM $\uparrow$ &
            GE $\uparrow$ & S $\uparrow$\\
        \midrule 
        \multirow{2}{*}[-0em]{ROME} & \textsc{GPT2-XL (1.5B)} & $100.0$ & $99.8$ & $97.9$ & $71.74$ & $75.31$ & $10.48$ & $618.6$ & $89.57$\\
        & \textsc{GPT-J (6B)}& $100.0$ & $99.8$ & $97.25$ & $73.65$ & $81.94$ & $13.92$ & $617.1$ & $92.34$ \\
        & \textsc{Llama-2 (7B)}& $100.0$ & $99.9$ & $96.7$ & $68.65$ & $80.79$ & $20.62$ & $585.96$ & $91.69$ \\
        \midrule
        
        \multirow{2}{*}[-0em]{MEMIT} & \textsc{GPT2-XL (1.5B)} & $100.0$ & $99.7$ & $97.85$ & $71.74$ & $75.21$ & $10.49$ & $618.54$ & $89.51$ \\
        & \textsc{GPT-J (6B)}& $100.0$ & $99.8$ & $97.05$ & $72.25$ & $82.06$ & $13.94$ & $616.6$ & $92.34$ \\
        & \textsc{Llama-2 (7B)}& $99.6$ & $97.4$ & $91.7$ & $57.8$ & $82.83$ & $21.68$ & $593.04$ & $90.86$ \\
        \midrule\bottomrule
    \end{tabular*}
        \caption{Comparison between ROME and MEMIT when editing only a single layer for CounterFact dataset.}\label{table:rome-vs-memit-counterfact}
    \vskip -0.0in
\end{table*}

The above optimization objective aims to modify the output representations of vectors in $K_E$ to $V_E$ by minimizing the least square error between them instead of requiring them to be equal with an equality constraint. This is the major difference between the objectives of ROME and MEMIT, where ROME poses the memorization part of the objective as an equality constraint whereas MEMIT relaxes the equality constraint to a least-square objective. This allows \citet{MEMIT} to find a closed-form solution for making $E$ edits to the model in a single update, represented by the matrix $K_E$. The solution for the MEMIT objective is:

\begin{equation}\label{eq:memit}
\begin{aligned}
    \hat{W} &= W_0 + \Delta \hspace{10pt} \text{where} \hspace{10pt}  
    \\ \Delta &= \big(V_E - W_0K_E \big) K_E^T \big( \lambda C_0 + K_EK_E^T \big)^{-1}
\end{aligned}
\end{equation}

We deliberately write the first term in both solutions in a similar form. The first term in $\Delta$ represents the residual error (represented by $R$) of the new associations ($K_E, V_E$) when evaluated on the old weights $W_0$. $R \triangleq v_e - W_0k_e$ is a vector in case of ROME since we are only able to make singular edits, whereas $R \triangleq V_E - W_0K_E$ is a matrix for MEMIT consisting of a row of vectors corresponding to each edit in the batch. 

To summarize, in this section we show that ROME and MEMIT can be seen as two realizations of the \textit{preservation-memorization} (PM) objective of model editing, where ROME enforces memorization using an equality constraint whereas MEMIT enforces memorization as a least square objective. The least-square constraint in MEMIT allows to reach a closed form solution for batch updates. 

\begin{figure*}
    \centering
    
    % First row with three subfigures
    \begin{subfigure}[b]{0.26\textwidth}
        \centering
        \includegraphics[width=\textwidth]{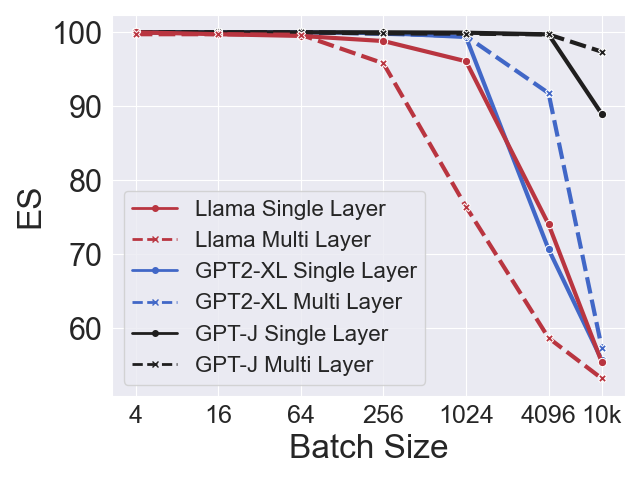}
        \caption{Efficacy Score (ES)}
        \label{fig:sub1}
    \end{subfigure}
    \hfill % This adds a space between the subfigures
    \begin{subfigure}[b]{0.26\textwidth}
        \centering
        \includegraphics[width=\textwidth]{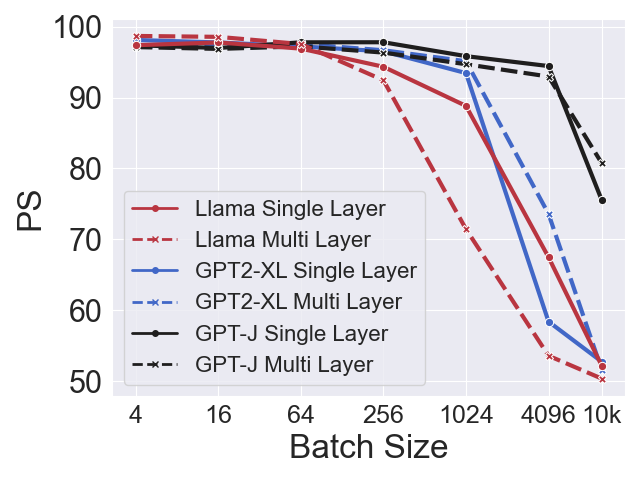}
        \caption{Paraphrase Score (PS)}
        \label{fig:sub2}
    \end{subfigure}
    \hfill % This adds a space between the subfigures
    \begin{subfigure}[b]{0.26\textwidth}
        \centering
        \includegraphics[width=\textwidth]{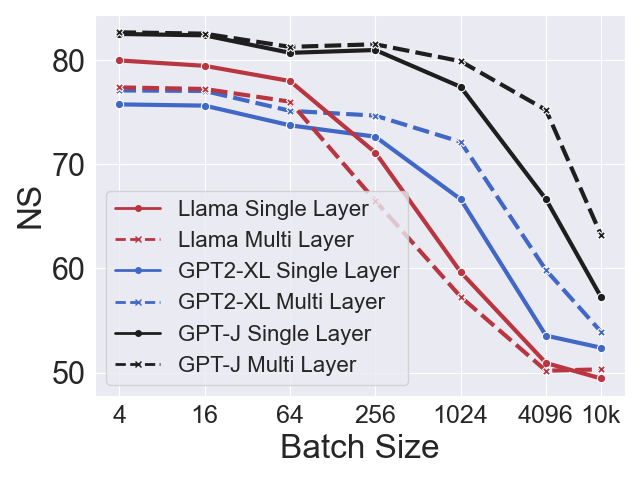}
        \caption{Neighborhood Score (NS)}
        \label{fig:sub3}
    \end{subfigure}
    
    \caption{Performance comparison of model editing using MEMIT when editing just one layer against multiple layers using the MEMIT edit-distribution algorithm on the CounterFact dataset.}
    \label{fig:memit-single-vs-multi}
\end{figure*}

\section{Edit-Distribution Algorithms} 
The difference in objectives is not the only difference between ROME and MEMIT. MEMIT \citep{MEMIT} also additionally distributes its edits into multiple layers, which has been one of the reasons for success of MEMIT at large batch sizes. This distribution is done by using the formula:

\begin{equation}
\Delta^l = \frac{\big(V^L_E - W_0^lK_E^l \big)}{L - l +1} {K^l}_E^T \big(C_0^l + K^l_E {K^l}_E^T \big)^{-1}
\end{equation}

% This is almost the same formula as equation \ref{eq:memit} with minor changes.
where $\Delta^l$ represents the change in weights at layer $l$, where $l \in \{L-(n-1), L-(n-2), \dots L\}$ represents one of the $n$ layers being edited. $V^L_E = V_E$ are the representations of the fact being edited at the final edit layer, which is represented by $L$. All other representations of $K_E$ and $C_0$ are calculated at the layer $l$ being edited. For $n=1$, the formula reduces to equation \ref{eq:memit}. We call this algorithm a type of \textbf{edit-distribution algorithm}, which is applied post-hoc after finding the closed-form solutions to the PM-objective. 

% In this paper, we want to disentangle these two different parts of MEMIT. 
The edit-distribution algorithm is separate from the solutions of the ROME and MEMIT objectives, therefore, we can apply the edit-distribution algorithm when using ROME, as well as use MEMIT without distributing the edits into multiple layers. The formula for using the MEMIT edit-distribution algorithm on ROME is as follows:

\begin{equation}
    \Delta^l = (v^L_e - W^l_0 k^l_e) \frac{k^{l^T}_e C^{l^{-1}}_0}{k^{l^T}_e C^{l^{-1}}_0 k^l_e}
\end{equation}

Prior works on model editing do not differentiate between the MEMIT-objective and the edit-distribution algorithm, and as a consequence we never see edits using ROME being distributed to multiple layers or MEMIT being used on only a single layer. The additional wrapping of edit-distribution also makes MEMIT seem distant from ROME. In the next section, we remove the wrapping of edit-distribution from MEMIT and allow for a fair comparison between the two algorithms. 
% The unified code for the two algorithms can be found \href{https://github.com/scalable-model-editing/unified-model-editing}{here}\footnote{\url{https://github.com/scalable-model-editing/unified-model-editing}}. With this code-base, the MEMIT edit-distribution algorithm can be applied to ROME as well as EMMET which we introduce later.

%\subsection{Single Layer Edits Using ROME and MEMIT}

%The results are shown in Table \ref{table:rome-vs-memit-counterfact}. We see that solutions to both ROME and MEMIT objectives perform equally well at making singular edits across different metrics, without needing to distribute the edits to multiple layers. The opposite test to this would be to make multi-layer edits using both ROME and MEMIT. Edit-distribution algorithms are usually applied to improve performance of model editing when similar performance cannot be reached by just editing one layer. Since the edit scores are close to perfect for singular edits, applying edit-distribution algorithms are not useful and produce similar results. This can be seen in Table \ref{table:rome-vs-memit-multi}.

% DO THE MULTI-LAYER EXPERIMENTS WITH MEMIT

\subsection{Impact of edit-distribution Algorithms}
%In the previous section, we saw that both ROME and MEMIT objectives perform equally well when making singular edits to the model. 

The key advantage of the edit-distribution algorithm is apparent when making batched edits. In this section, we perform two experiments to analyze this. First, we compare single edits in ROME and MEMIT with and without edit distribution on 1k randomly selected facts from the CounterFact datase \cite{ROME}. Following that, we compare batched editing in MEMIT with and without edit distribution. Both experiments are performed on three different models - GPT2-XL (1.5B) \cite{gpt-2}, GPT-J (6B) \cite{gpt-j} and Llama2-7B \cite{llama2}. %The implementation details about selected layers and hyperparameters are provided in section \ref{appendix:implementation-details}. 

%In this section, we evaluate the performance of making batched edits using MEMIT as a function of the batch size on the CounterFact dataset. 

The results are shown in Table \ref{table:rome-vs-memit-counterfact} for edits without distribution and Table \ref{table:rome-vs-memit-multi} (appendix) for edits with distribution. We use the more stable version of ROME called r-ROME as presented in \citep{akshat-rebuilding} that does not lead to model collapse and improves generalization. We see that solutions to both ROME and MEMIT objectives perform equally well at making singular edits across different metrics, without needing to distribute the edits to multiple layers. 
To highlight the usefulness of edit-distribution algorithms, we make batched edits with MEMIT comparing performance with and without edit distribution. The results are shown in Figure \ref{fig:memit-single-vs-multi}. When only editing a single layer, we see that MEMIT is able to successfully make batched edits up to a batch size of 1024 for GPT2-XL, 256 for Llama-2-7b and a batch-size as large as 4096 for GPT-J\footnote{In our experiments we find GPT-J to be an easier model to edit compared to other models. This is both intriguing but also not the best model to evaluate model editing success.}. After this point, the performance of model editing increases when making edits on multiple layers, except for Llama-2-7b. All hyperparameters for all models were chosen as is from prior work \citep{ROME, MEMIT, editing-survey, survey-comprehensive} (appendix \ref{appendix:implementation-details}).

%Edit-distribution algorithms are usually applied to improve performance of model editing when similar performance cannot be reached by just editing one layer. Since the edit scores are close to perfect for singular edits, applying edit-distribution algorithms are not useful and produce similar results. 

%For GPT2-XL, we see that the edit score improves for batch size greater than 1024 with edit-distribution algorithms. The same is true for GPT-J beyond a batch size of 4096. If we consider all metrics together, as done in the "Score" metric, we can see improvements with edit-distribution starting at batch size 256 for GPT2-XL and GPT-J. This shows that edit-distribution algorithms become important as the batch size of the edits increase.

%Throughout our experiments, we found that ROME and MEMIT perform well on GPT-J. However, GPT-J may be an easier model to edit. To investigate this, we also performed experiments on Llama-2-7b, selecting layers based on prior works\citep{editing-survey, survey-comprehensive}. We see that edit-distribution using the algorithm proposed in \cite{MEMIT} hurts the model editing performance for Llama-2-7b, as shown in Figure \ref{fig:memit-single-vs-multi}. 

With these experiments, we want to highlight two key points - firstly, when comparing the effectiveness of two optimization objectives, the evaluation should not be conflated with the edit distribution algorithms. After removing the wrapping of edit-distribution from MEMIT, we see that the performance numbers for ROME and MEMIT have an uncanny similarity. Secondly, the MEMIT edit-distribution algorithm is not perfect and currently is the only way to distribute edits into multiple layers, where the residual in the update is distributed with specific ratios through different layers. We hope these experiments will bring more focus to edit distribution algorithms and boost further research in these methods.

%This highlights two key points: 1) edit-distribution algorithms are separate from the optimization objective of MEMIT, and can be used with any model editing algorithms, and 2) the MEMIT edit-distribution algorithm is not perfect, as we see in loss of performance when using Llama-2-7b in Figure \ref{fig:memit-single-vs-multi}. The residual is distributed in a very specific way in multiple layers, where the residual $R = \big(V^L_E - W_0^lK_E^l \big)/(L - l +1)$ is divided by the ratio of how far the edited layer is from the final layer being edited. There are possibly many other unexplored ways of distributing edits to multiple layers. With these experiment, we want to underscores the need for further research into edit-distribution algorithms, as it is a critical component for effective batch-wise model editing.

\begin{figure*}
    \centering
    
    % First row with three subfigures
    \begin{subfigure}[b]{0.27\textwidth}
        \centering
        \includegraphics[width=\textwidth]{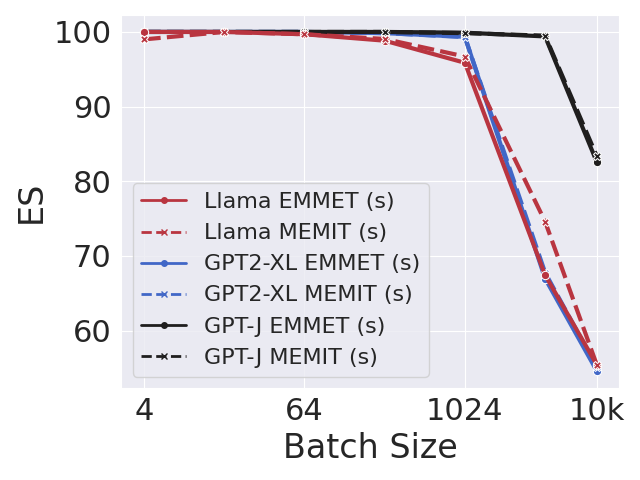}
        \caption{Efficacy Score (ES)}
        \label{fig:sub1}
    \end{subfigure}
    \hfill % This adds a space between the subfigures
    \begin{subfigure}[b]{0.27\textwidth}
        \centering
        \includegraphics[width=\textwidth]{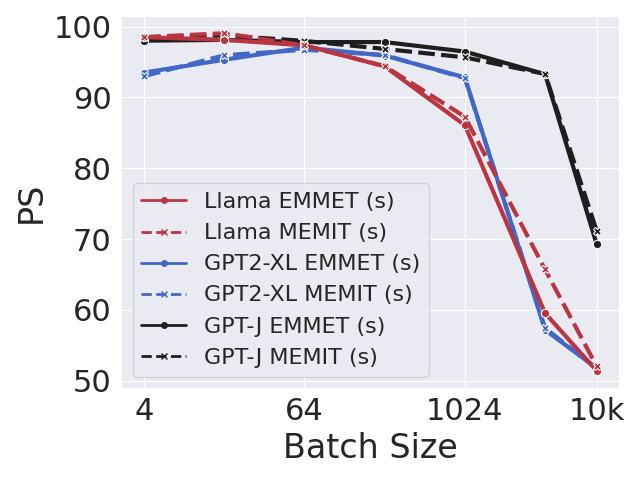}
        \caption{Paraphrase Score (PS)}
        \label{fig:sub2}
    \end{subfigure}
    \hfill % This adds a space between the subfigures
    \begin{subfigure}[b]{0.27\textwidth}
        \centering
        \includegraphics[width=\textwidth]{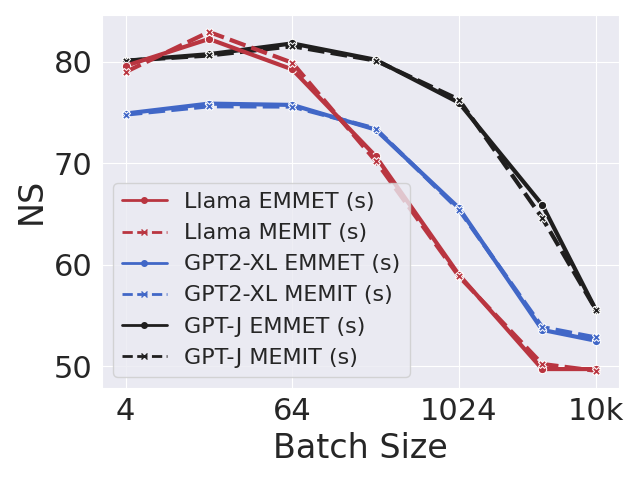}
        \caption{Neighborhood Score (NS)}
        \label{fig:sub3}
    \end{subfigure}
    
    \vspace{1em} % Adds some vertical space between the rows
    \hspace{20mm}
    % Second row with two subfigures
    \begin{subfigure}[b]{0.27\textwidth}
        \centering
        \includegraphics[width=\textwidth]{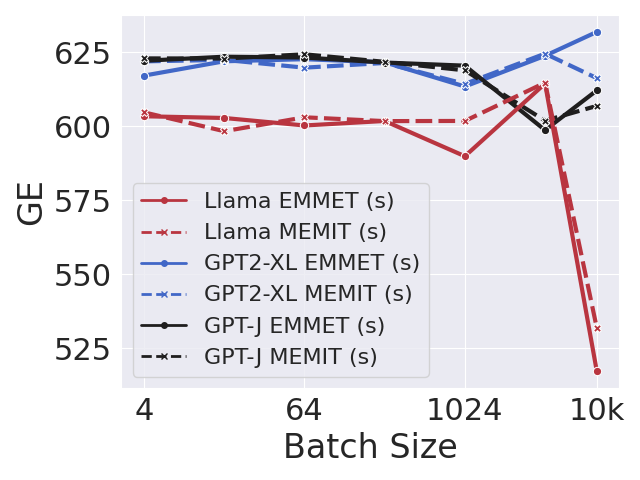}
        \caption{Generation Entropy (GE)}
        \label{fig:sub4}
    \end{subfigure}
    \hfill % This adds a space between the subfigures
    \begin{subfigure}[b]{0.27\textwidth}
        \centering
        \includegraphics[width=\textwidth]{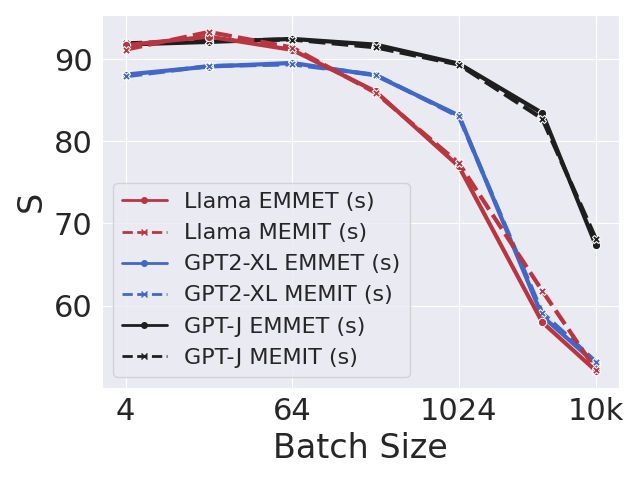}
        \caption{Score (S)}
        \label{fig:sub5}
    \end{subfigure}
    \hspace{20mm}
    
    \caption{Single layer editing performance of EMMET as a function of batch size when compared to MEMIT on the CounterFact dataset.}
    \label{fig:emmet-vs-memit-single}
\end{figure*}

\section{Introducing EMMET}
In section \ref{sec:pm}, we show that ROME and MEMIT are both algorithms optimizing the preservation-memorization objective of model editing, where ROME does memorization using an equality constraint wherease MEMIT uses a least-square objective for memorization. Thus, we ask the question - \textit{can we perform batched-editing under an equality constraint for memorization?} 

In this section, we provide a closed-form solution for batched-editing where memorization is done with equality constraints under the presevation-memorization objective, and thus present a batched-version of ROME, a method we call \textbf{EMMET} - \textbf{\underline{E}}quality-constrained \textbf{\underline{M}}ass \textbf{\underline{M}}odel \textbf{\underline{E}}diting in a \textbf{\underline{T}}ransformer.

Let $K_0 = [k^0_1 \hspace{4pt}| k^0_2 \hspace{4pt}| \dots |\hspace{4pt}k^0_N]$ represent $N$ key-vectors whose representations we want to preserve. Additionally, let $k^e_1, k^e_2 \dots k^e_E$ represent key-vectors for $E$ facts we want to edit in the model at the same time. Then according to the preservation-memorization objective, we want to find new weights $\hat{W}$ for a weight matrix $W_0$ such that:

\begin{equation}
\begin{aligned}
    \underset{\hat{W}}{\operatorname{argmin}} \hspace{4pt} \underbrace{\left\| \hat{W} K_0 - W_0 K_0 \right\|^2_F}_{\text{preservation}} \hspace{8pt}  \text{    s.t. } \\ \hspace{8pt} \underbrace{\hat{W} k^e_i = v^e_i \hspace{8pt} \forall i \in [1, 2 \dots E]}_{\text{memorization}}
\end{aligned}
\end{equation}

As can be seen in the above equation, the preservation of representations happens in the first term whereas memorization of all the new facts are forced using an equality constraint in the second term. The above equation is solved using lagrange-multipliers. The derivation of the above equation for the generalized case of batched editing can be found in Appendix \ref{sec:derivation}.

\begin{figure*}
    \centering
    
    % First row with three subfigures
    \begin{subfigure}[b]{0.27\textwidth}
        \centering
        \includegraphics[width=\textwidth]{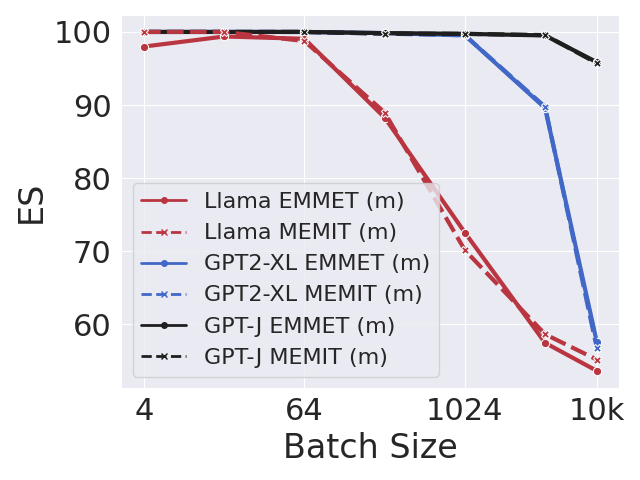}
        \caption{Efficacy Score (ES)}
        \label{fig:sub1}
    \end{subfigure}
    \hfill % This adds a space between the subfigures
    \begin{subfigure}[b]{0.27\textwidth}
        \centering
        \includegraphics[width=\textwidth]{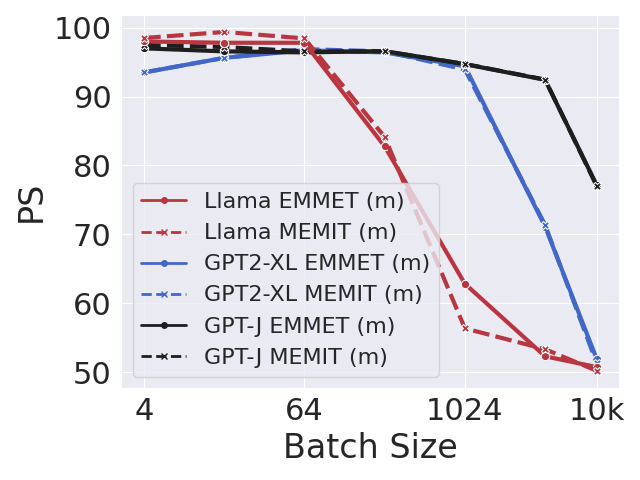}
        \caption{Paraphrase Score (PS)}
        \label{fig:sub2}
    \end{subfigure}
    \hfill % This adds a space between the subfigures
    \begin{subfigure}[b]{0.27\textwidth}
        \centering
        \includegraphics[width=\textwidth]{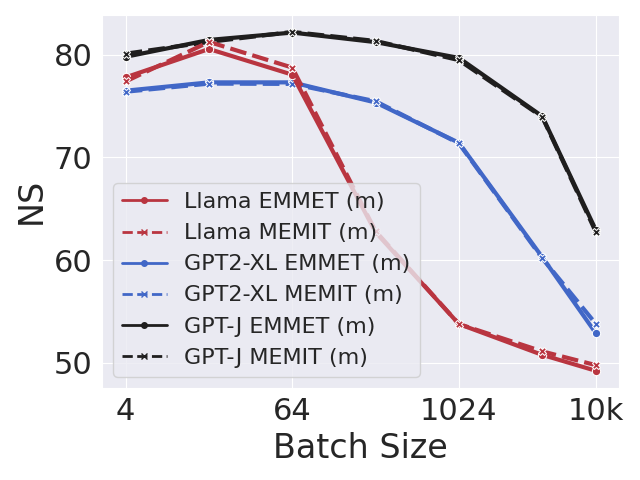}
        \caption{Neighborhood Score (NS)}
        \label{fig:sub3}
    \end{subfigure}
    
    \vspace{1em} % Adds some vertical space between the rows
    \hspace{20mm}
    % Second row with two subfigures
    \begin{subfigure}[b]{0.27\textwidth}
        \centering
        \includegraphics[width=\textwidth]{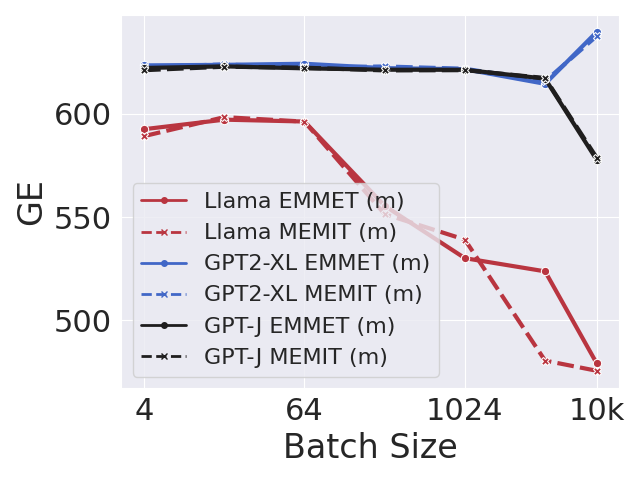}
        \caption{Generation Entropy (GE)}
        \label{fig:sub4}
    \end{subfigure}
    \hfill % This adds a space between the subfigures
    \begin{subfigure}[b]{0.27\textwidth}
        \centering
        \includegraphics[width=\textwidth]{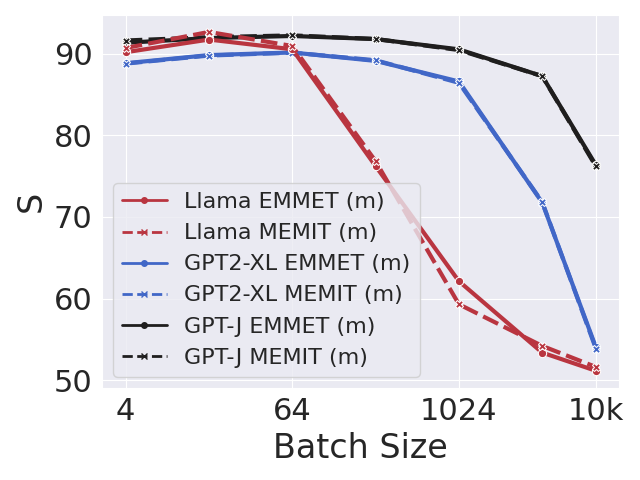}
        \caption{Score (S)}
        \label{fig:sub5}
    \end{subfigure}
    \hspace{20mm}
    
    \caption{Performance comparison of EMMET and MEMIT when distributing the edit over multiple layers using the MEMIT edit-distribution algorithm on the CounterFact dataset.}
    \label{fig:emmet-vs-memit-multi}
\end{figure*}

The closed form solution for batched editing with equality-constraint or EMMET is shown below:

\begin{equation}\label{eq:emmet_main}
\begin{aligned}
 &\hat{W} = W_0 + \Delta \hspace{10pt} \text{where} 
 \\ \Delta &= \left ( V_E - W_0 K_E \right ) \left ( K_E^T C^{-1}_0 K_E \right )^{-1} K_E^TC^{-1}_0 
 \end{aligned}
\end{equation}

Here, $C_0 = K_0K_0^T$ has the usual meaning as in the derivation of ROME and MEMIT, where $K_0$ contains the list of representations we want preserved during editing. We write the update equation for EMMET in a familiar form, where the residual $R =  V_E - W_0 K_E$ is modified by some matrix operations to update the models with new edits. Additionally, when we put $E=1$, the $K_E$ matrix reduces to a single vector $k_e$ and equation \ref{eq:emmet_main} reduces to the ROME update equation (equation \ref{eq:rome-update-equation}). With EMMET, we complete the unification of ROME and MEMIT under the preservation-memorization objective and achieve a symmetry with the usage of these algorithms. EMMET allows for making batched-edits as well as singular when using equality constraints for memorization, much similar to MEMIT with least-square based memorization. %The invertibility of matrix $B = K_E^T C^{-1}_0 K_E$ cannot be guaranteed just like $C_0$, but we find that in practice, $B$ is usually invertible for smaller batch sizes like $C_0$.

\subsection{Stabilizing EMMET}
There are two important matrices that are being inverted in EMMET and MEMIT. The first one is $C_0 = K_0 K^T_0$, which is defined identically in both algorithms, whereas $D = K_E^T C^{-1}_0 K_E$ is only inverted in EMMET. While the invertibility of both matrices are assumed, they are not always guaranteed. Each of the matrices $K_0$ or $K_E$ can be written as a row of column vectors as explained in section \ref{sec:pm}, and thus $C_0$ can be written as a sum of outer products: 

\begin{equation}
    C_0 = K_0 K^T_0 = \sum_i k^0_ik^{0^T}_i
\end{equation}

where $k^0_i$ represents a key-vector we want to preserve. For an LLM of dimension $d$, the dimensionality of a key-vector is usually $4d$ (Figure \ref{fig:diagram}), which is the dimensionality of the square matrix $C_0$. If $C_0$ is a $4d$-dimensional square matrix which is a summation of rank-1 matrices, it is invertible as long as there are atleast $4d$-independent vectors in the summation, or $4d$-independent vectors in $K_0$. For example, for GPT2-XL with hidden dimension of 1600, the dimensionality of key vectors are 6400. So as long as representations of atleast 6400 independent key-vectors are being preserved while editing, $C_0$ will be an invertible matrix. In practice, we preserve representations of a much larger number of vectors, and hence this condition is always satisfied.

The matrix $D = K_E^T C^{-1}_0 K_E$ is a square matrix of dimensionality equal to the number of edits. If given that $C_0$ is invertible, $D$ is invertible as long as $K_E$ is full-rank, which means all key-vectors corresponding to facts being memorized are independent of each other. While this is not guaranteed, it can be verified before editing and facts corresponding to non-independent keys can be removed from a batch. In practice, we do not find invertibility of D being an issue. However, we find that $D$ is often ill-conditioned, which means that the ratio of the largest and smallest eigenvalues of $D$ explodes. This doesn't necessarily mean that the matrix is singular (non-invertible), but it does mean that numerical computations involving the matrix inverse are unstable and can lead to large numerical errors. To counter this, we set $D = D + \alpha I$, where $\alpha$ is set to 0.1 after an ablation over multiple batch sizes. This allows for stable batched edits using EMMET and also ensures that the $D$ matrix is always invertible.

\subsection{Batch Editing with EMMET}
We begin by experimenting with EMMET for model editing with varied batch sizes on GPT2-XL, GPT-J and Llama-2-7b on the CounterFact and zsRE \cite{zsre} datasets. The exact implementation details can be found in section \ref{appendix:implementation-details}. We compare the performance of EMMET and MEMIT on batch sizes up to 10,000 while editing both single (to directly compare the optimization objectives) and multiple layers. The single layer editing comparison between EMMET and MEMIT can be found in Figure \ref{fig:emmet-vs-memit-single}. We see that both methods have almost identical performance in practice across different metrics. MEMIT performs slightly better than EMMET for Llama-2-7b, as indicated by ES, PS and S metrics. We then apply the MEMIT edit-distribution on EMMET and compare it with MEMIT. The results are shown in Figure \ref{fig:emmet-vs-memit-multi}. We see that in this case, EMMET performs slightly better than MEMIT for Llama-2-7b. The results on the zsRE dataset tell a similar story and can be seen in Figure \ref{fig:emmet-zsre-single} and \ref{fig:emmet-zsre-multi}. The experiments for different hyperparameter values are shown in Appendix \ref{sec:appendix-hparams}. These results present EMMET as a viable new batched-editing algorithm.

%We show that EMMET can perform batched edits with large batch sizes and performs comparably to MEMIT. 
%As shown previously, we disentangle the optimization objective with edit-distribution algorithms and show the comparison between EMMET and MEMIT when editing a single layer. We use identical configurations for both models including the layers being edited. 

\begin{figure}
    \centering
    \begin{subfigure}{.24\textwidth}
        \centering
        \includegraphics[width=\linewidth]{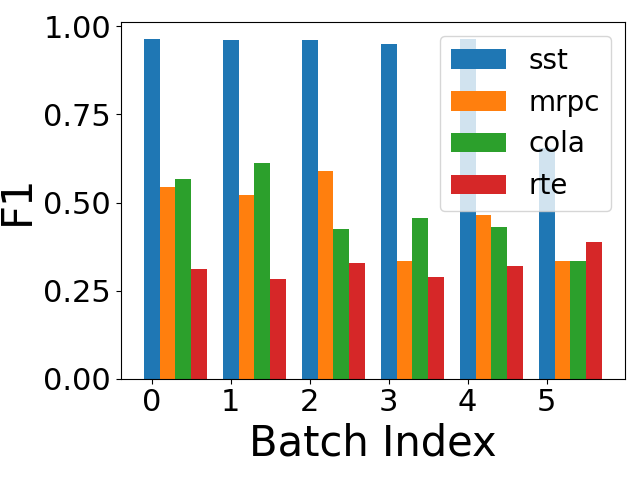}
        \caption{EMMET}
        \label{fig:memit_gptj:edit_score}
    \end{subfigure}%
    \begin{subfigure}{.24\textwidth}
        \centering
        \includegraphics[width=\linewidth]{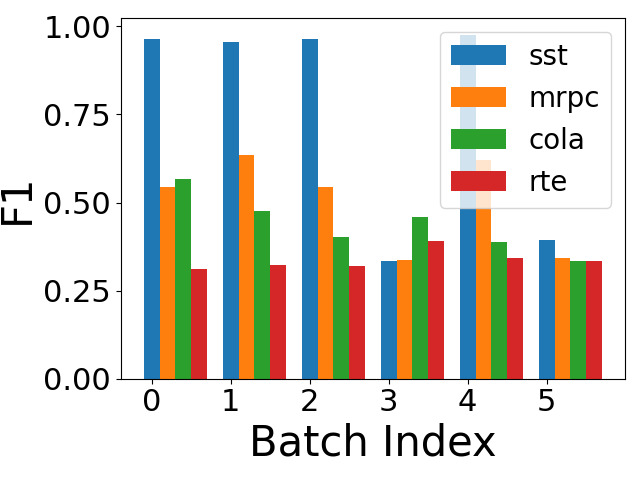}
        \caption{MEMIT}
        \label{fig:memit_gptj:downstream}
    \end{subfigure}
    \caption{Downstream performance of post-edit Llama2-7b model for EMMET and MEMIT on four GLUE tasks. Batch index 0 refers to downstream performance before editing, with the performance of 5 independent edits of batch size 256. }
    \label{fig:downstream_256}
\end{figure}

Previous work \citep{hurt, akshat-catastrophic} has shown that model editing is often accompanied by model degradation. This was shown by evaluating the edited model on downstream tasks from the popular GLUE benchmark \cite{glue}. Once we identified that memorization in MEMIT is happening using an approximate least-square constraint rather than an equality constraint, we hypothesised that a possible reason for model degradation could be the use of the least-square constraint. Thus, using an equality constraint, which by definition requires the edit to be exact, may not degrade other knowledge or skills of the model. This was also the motivation behind generalizing ROME to batched edits in the form of EMMET. To test this hypothesis, we adopt the evaluation setting of \citet{akshat-catastrophic} and evaluate both EMMET and MEMIT on four downstream tasks - sentiment analysis (SST2) \cite{sst2}, paraphrase detection (MRPC) \cite{mrpc}, natural language inference (NLI) \citep{nli1, nli2, nli3, nli4} and linguistic acceptability classification \cite{cola} for doing downstream evaluation. The results are shown in Figure \ref{fig:downstream_256} for a batch size of 256. The results for other batch sizes can be found in Appendix \ref{appendix:implementation-details}. We find that both EMMET and MEMIT also degrade the model similarly. 

The fact that both EMMET and MEMIT perform editing and degrade the model with an uncanny similarity shows that a "stronger" equality constraint does not enable more accurate model editing. We believe reason behind this is the construction of the key-vector, which is created by taking the average of  representations of multiple phrasings of a fact (appendix \ref{sec:key-value-creation}). This is done to make edits that generalize beyond a single phrasing of a fact. As the key-vector is an averaged representation over randomly selected phrasings, it is an approximation of the ideal vector representation of a fact. We believe that such an approximate representation does not require the additional accuracy of memorization enforced due to the equality constraint. Our findings also indicate that we may be reaching the limit of model editing capabilities under the preservation-memorization objective.

\subsection{Related Work}\label{sec:related work}
Model editing methods can be broadly classified into two types - methods that add information in-context \citep{SERAC, mquake, ripple-effects}, and methods that modify the parameters of underlying model \citep{metamodel, MEND, ROME, MEMIT, MALMEN}. Various model editing techniques have been proposed in the past that tackle this problem in different ways. \cite{knowledgeneurons} first identify knowledge containing neurons in a model using integrated gradients \citep{integrated-gradients} and then modify the selected neurons to edit facts in a model. This method is not scalable with increasing model sizes as it requires us to find activations for each neuron in the model. \cite{metamodel} and \cite{MEND} train a hypernetwork \citep{hypernetwork} that generates the new weights of the model being edited. While these methods have been optimized to scale with a square-root dependence on the size of the edited model, it still requires training of additional editing models dependent on each source model being edited. Other methods add the most relevant updated knowledge in context \citep{SERAC, ripple-effects, mquake}. While such methods provide a viable alternative to model editing, in this paper, we focus on parameter-modifying model editing methods, namely ROME \cite{ROME} and \cite{MEMIT}.

\section{Conclusion}
In this paper we unite two popular model editing techniques, ROME and MEMIT, under the \textbf{preservation-memorization} objective, with ROME performing equality-constrained edits and MEMIT operating under a least-square constraint. We disentangle the \textit{edit-distribution} algorithm proposed in MEMIT from the optimization objective, presenting them as separate entities. We also present EMMET, a new batched-editing algorithm based on the preservation-memorization objective, where memorization happens under an equality constraint. Our experiments show that EMMET has similar performance to MEMIT across multiple dimensions and metrics. 

Enabling batched editing with equality-constraint in the form of EMMET allows us to truly unify ROME and MEMIT and shows that both these algorithms are essentially equivalent in terms of their (i) optimization objective, (ii) their abilities (singular and batched editing, a symmetry enabled by EMMET), (iii) their model editing performance and (iv) their limitations (similar model degradation). \textbf{EMMET is a cornerstone in completing this larger picture.} These results suggest that EMMET (or ROME) and MEMIT not only have very similar theoretical roots but also perform similarly in practice. The unified framework presented in our work along with the disentanglement of edit distribution algorithm has also enabled a fair comparison between the two algorithms, which was not possible before our work. We hope that this framework facilitates ease of comparison, consistency of implementation, and a much deeper understanding of these model editing methods.

 %emphasizing that a fair comparison of future model editing techniques with MEMIT should be based on the objective of MEMIT rather than conflating it with the edit-distribution algorithm. %This allows researchers to take advantage of the general benefits of having a unified framework such as ease of comparison (this is a major benefit in our opinion - A fair comparison between equality constraints and least squares constraints for model editing was not possible before our work), consistency of implementation, and ease of understanding the two methods

%EMMET completes batched editing using both types of objectives and truly unifies model editing under the preservation memorization framework. We hope that this unifying framework improves the intuitive understanding of these algorithms and fuels future research based on both intuition and mathematics. 
%\bibliographystyle{abbrv}

\section{Limitations}
While our technique may streamline error correction processes, it does not address deeper structural limitations within models, such as edited models inadvertently amplifying existing errors or introducing new inaccuracies. Furthermore, the effectiveness of our method varies depending on the complexity of the model architecture and the nature of the edited knowledge as evidenced by our experiments. Despite having a theoretically `stronger' memorization objective, EMMET is not able to outperform MEMIT, which also indicates that we might have reached a saturation point for model editing using naive implementations of the preservation-memorization objective, underscoring the fact that significant progress is yet to be made in understanding edit distribution and its implications. %Thus, while our work contributes to a deeper understanding of model behavior, it is essential to recognize and account for these limitations in the interpretation and application of our findings.

\section{Ethical Considerations}
While our model editing method allows users to effectively correct for errors or update facts in models, caution is warranted. Our technique also introduces concerns for potential misuse such as malicious actors inserting harmful or false knowledge in LLMs that is absent from the original training data. As such, we warn readers that LLMs should not be considered reliable knowledge bases. 

\bibstyle{plainnat}
% \bibliography{model_editing}

\bibliography{custom}

\appendix
\section{Appendix}

\begin{table}[h]
    \centering
    \begin{tabular}{c|c|c} % l = left, c = center, r = right

    \textbf{Batch Size} & \textbf{Num Batches} & \textbf{Total Edits} \\ \hline
    4 & 25 & 100 \\ \hline
    16 & 10 & 160 \\ \hline
    64 & 5 & 320 \\ \hline
    256 & 5 & 1280 \\ \hline
    1024 & 3 & 3072 \\ \hline
    4096 & 2 & 8192 \\ \hline
    10,000 & 1 & 10,000 \\ \hline
    \end{tabular}
    \caption{Statistics for batch size and number of batches used to create the numbers for this paper.}
    \label{tab:batch-size}
\end{table}

\begin{figure*}
    \centering
    
    % First row with three subfigures
    \begin{subfigure}[b]{0.32\textwidth}
        \centering
        \includegraphics[width=\textwidth]{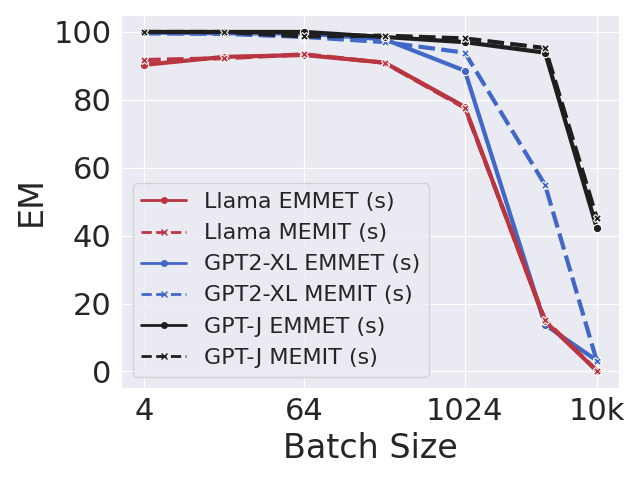}
        \caption{Efficacy Accuracy (EM)}
        \label{fig:sub1}
    \end{subfigure}
    \hfill % This adds a space between the subfigures
    \begin{subfigure}[b]{0.32\textwidth}
        \centering
        \includegraphics[width=\textwidth]{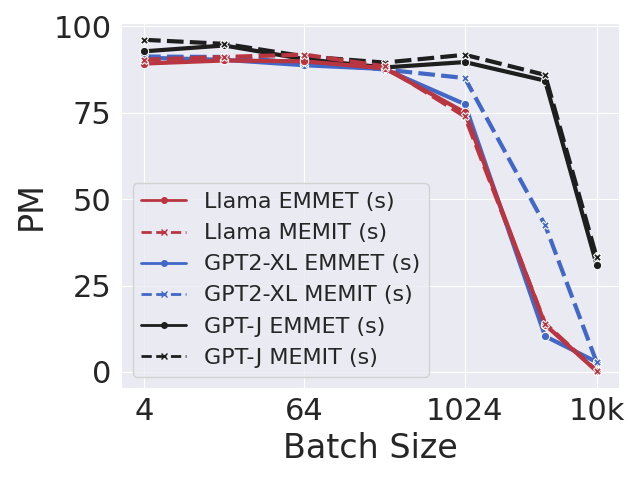}
        \caption{Paraphrase Accuracy (PM)}
        \label{fig:sub2}
    \end{subfigure}
    \hfill % This adds a space between the subfigures
    \begin{subfigure}[b]{0.32\textwidth}
        \centering
        \includegraphics[width=\textwidth]{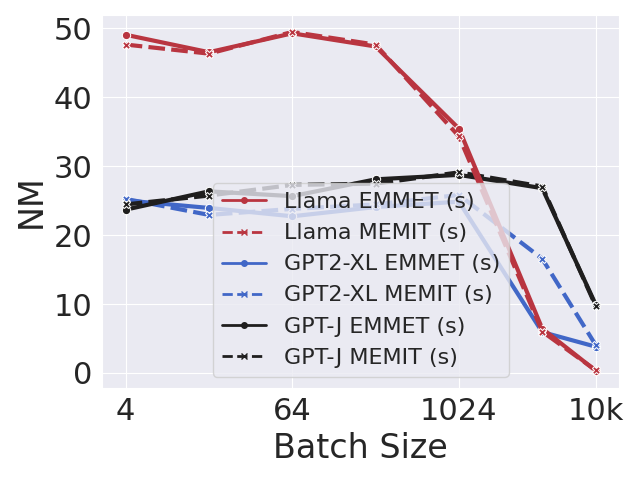}
        \caption{Neighborhood Accuracy (NM)}
        \label{fig:sub3}
    \end{subfigure}
    
    \caption{Single layer editing performance of EMMET as a function of batch size when compared to MEMIT on the zsRE dataset.}
    \label{fig:emmet-zsre-single}
\end{figure*}

\begin{table*}
    \scriptsize
    % \small
    % {\textwidth}{@{\extracolsep{\fill}}*{11}{c}}
    \setlength\tabcolsep{0pt}
    \setlength\extrarowheight{1pt}
    % \begin{tabular}{c c c c c c c c c c c}
    \begin{tabular*}{\textwidth}{@{\extracolsep{\fill\centering}}*{10}{c}}
        \toprule 
        % \multicolumn{12}{c}{DATASET: MPI-1k $\vert$ MODEL: \textsc{OPT}}\\
        % \midrule
        \multirow{2}{*}[-0.3em]{\textsc{Algorithm}} & 
            \multirow{2}{*}[-0.3em]{\textsc{Model}} &
            \multicolumn{2}{c}{Efficacy} & 
            \multicolumn{2}{c}{Generalization} & 
            \multicolumn{2}{c}{Locality} & 
            \multicolumn{1}{c}{Fluency} & 
            \multicolumn{1}{c}{Score}\\ 
        \addlinespace[0.125em] \cline{3-10} \addlinespace[0.25em]
        &  & ES $\uparrow$ & EM $\uparrow$ &
             PS $\uparrow$ & PM $\uparrow$ &
             NS $\uparrow$ & NM $\uparrow$ &
            GE $\uparrow$ & S $\uparrow$\\
        \midrule 
        \multirow{2}{*}[-0em]{ROME} & \textsc{GPT2-XL (1.5B)} & $100.0$ & $99.79$ & $97.78$ & $71.75$ & $76.16$ & $10.93$ & $617.56$ & $89.93$\\
        & \textsc{GPT-J (6B)}& $100.0$ & $99.8$ & $97.95$ & $72.07$ & $81.46$ & $13.42$ & $615.9$ & $92.35$ \\
        & \textsc{Llama-2 (7B)}& $99.68$ & $92.29$ & $98.1$ & $73.34$ & $77.59$ & $19.07$ & $589.44$ & $90.6$ \\
        \midrule
        
        \multirow{2}{*}[-0em]{MEMIT} & \textsc{GPT2-XL (1.5B)} & $100.0$ & $99.79$ & $97.57$ & $71.75$ & $76.14$ & $10.96$ & $617.9$ & $89.87$ \\
        & \textsc{GPT-J (6B)}& $100.0$ & $99.79$ & $97.1$ & $72.86$ & $81.96$ & $14.24$ & $615.97$ & $92.31$ \\
        & \textsc{Llama-2 (7B)}& $99.58$ & $91.34$ & $97.99$ & $72.18$ & $77.8$ & $19.27$ & $589.39$ & $90.63$ \\
        \midrule\bottomrule
    \end{tabular*}
        \caption{Comparison between ROME and MEMIT when editing multiple layers for the CounterFact dataset.}\label{table:rome-vs-memit-multi}
\end{table*}

\subsection{Implementation Details for ROME, MEMIT and EMMET}\label{appendix:implementation-details}
We use the standard implementation of ROME and MEMIT based on \cite{ROME} and \cite{MEMIT}. The range of layers edited for GPT2-XL is $[13,17]$ \cite{MEMIT}, for GPT-J is $[3-8]$ \cite{MEMIT} and for Llama-2-7b is $[4-8]$ \citep{editing-survey, survey-comprehensive}. In single layer editing experiments, layer 17 was edited for GPT2-XL \cite{ROME}, layer 5 was edited for GPT-J \cite{ROME}, and layer 5 was edited for Llama-2-7b \cite{editing-survey, survey-comprehensive}. These choices are directly taken from \cite{ROME} and \cite{MEMIT} for GPT2-XL and GPT-J. We follow the work of \cite{editing-survey} for choices of layers and hyperparameters for llama-2-7b. 

\begin{figure*}
    \centering
    
    \begin{subfigure}[b]{0.32\textwidth}
        \centering
        \includegraphics[width=\textwidth]{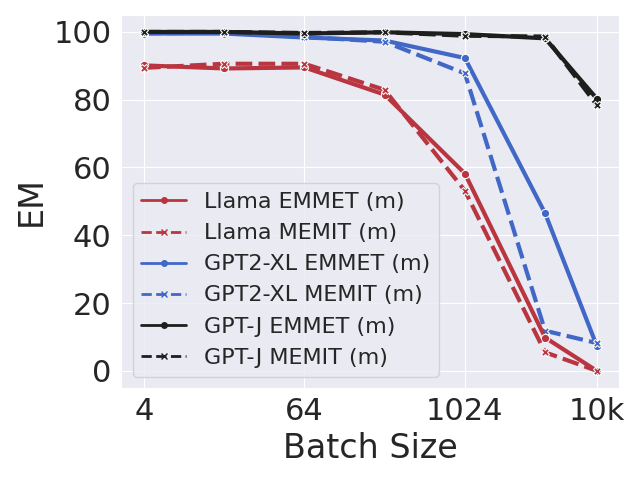}
        \caption{Efficacy Accuracy (EM)}
        \label{fig:sub1}
    \end{subfigure}
    \hfill % This adds a space between the subfigures
    \begin{subfigure}[b]{0.32\textwidth}
        \centering
        \includegraphics[width=\textwidth]{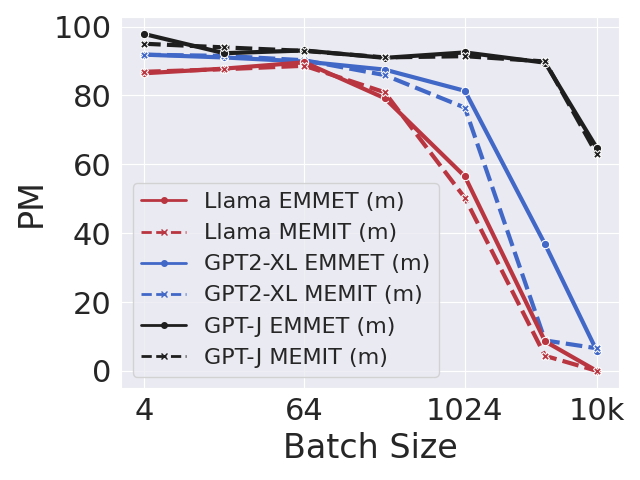}
        \caption{Paraphrase Accuracy (PM)}
        \label{fig:sub2}
    \end{subfigure}
    \hfill % This adds a space between the subfigures
    \begin{subfigure}[b]{0.32\textwidth}
        \centering
        \includegraphics[width=\textwidth]{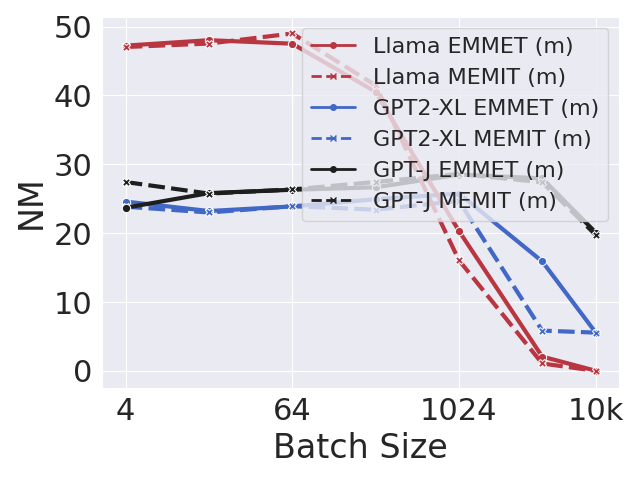}
        \caption{Neighborhood Accuracy (NM)}
        \label{fig:sub3}
    \end{subfigure}
    
    \caption{Multi layer editing performance of EMMET as a function of batch size when compared to MEMIT on the zsRE dataset.}
    \label{fig:emmet-zsre-multi}
\end{figure*}

We use the multi-counterfact dataset proposed in \citet{MEMIT} which is created by removing conflicting facts from the counterfact dataset \cite{ROME}. We then select a random sample of 10,000 facts so that the edits are influenced by the order in which the examples are presented in the dataset. To create the batched editing plots, we create multiple samples for each batch size and average over all the edits made in that set. We use batch sizes of 4, 16, 64, 256, 1024, 4096 and 10k. For each batch size, we use multiple batches and average the evaluation over the total number of batches. The statistics are shown in Table \ref{tab:batch-size}. For example, for a batch size of 1024, we first create 3 batches without replacement of size 1024, and perform batched edits on the 3 batches. The numbers are then reported by averaging the performance over 3*1024 facts which were edited in the model. We sample over a few batches so the results are not biased towards a single edited batched. We decrease the number of batches used in the sample due to computational reasons, as the amount of time for each experiment increases with the batch size. The same steps are followed for the zsRE dataset.

\subsection{Key-Value creation in ROME/MEMIT}\label{sec:key-value-creation}
We create key and value vectors for editing using the subject, relation, object framework presented in ROME \cite{ROME}. 

Sample queries under this formulation include: 
\vspace{10pt}

\scalebox{0.9}{
\begin{tabular}{c|c|c} % l = left, c = center, r = right

    \textbf{Subject} & \textbf{Prompt} & \textbf{Object} \\ \hline
    France & "The capital of \{S\} is \{O\}" & Paris \\ \hline
\end{tabular}
}
\vspace{10pt}

Model editing involves manipulating the model such that we're able to alter the object that is associated with a given input subject and prompt. In the table provided, the transformation from "Paris" to "London" exemplifies a potential application of model editing under the (s, r, o) formalization.

The subject and prompt together represent the key vector, which is found by averaging over a set of texts that end with the subject \emph{s} in the prompt \emph{p}: 

\begin{equation}
\begin{aligned}
    k_e = \frac{1}{N} \sum_{j=1}^{N} k(x_j + p) 
    \\ \text{where } k(x) = NL(W_{fc} a(x) + b_{fc}) 
    \\ \text{and }  a(x) = LN (\texttt{Att} (h^{l - 1}(x) ) + h^{l - 1} (x))
\end{aligned}
\end{equation}

$p$ is the prompt containing the subject and relation, and $x_j$ are 50 generated random sequences with lengths varying from 2 to 10 tokens to make the representation of the key vector more robust to paraphrasing. This also ensures that key vectors for different prompts are distinct enough as two base key vectors (with no random prefix) that have very similar representations move further apart when their representations with a prefix are averaged. LN represents layer normalization and NL is the non-linearity applied to the stream. 

Next, we choose a $v_e$ vector such that the new object $o^*$ is output for our $k_e$ vector. We set $v_e$ to minimize the loss as shown: 

\begin{equation}
    \begin{aligned}
        \underset{v_e}{\operatorname{argmin}} 
        \hspace{10pt}
        \frac{1} {N} \sum_{j = 1}^{N} -\log \mathbb{P}_{G(h^l = v_e)} [o^* \hspace{5pt} | 
        \hspace{5pt} x_j + p] 
        \\ + D_{KL} \left ( \mathbb{P}_{G(h^l = v_e)} [x \hspace{5pt}| \hspace{5pt}p'] \hspace{6pt} ||  \hspace{6pt} \mathbb{P}_{G(h^l) = v_e} [x \hspace{5pt}|\hspace{5pt} p'] \right ) 
    \end{aligned}
\end{equation}

The first term tries to maximize the probability of the target objective $o^*$ for a prompt of the form $x_j + p$ where $p$ is once again our desired prompt that was also used to generate the key vector. $G (v)$ represents the output of generation s.t. the hidden layer $h^l = v$. The second term tries to minimize the KL divergence when an unrelated prompt $p'$ is input to the model since we want our edit to keep unrelated knowledge unchanged.

We refer readers to the original ROME paper for more details on how key and value vector pairs $(k_e, v_e)$ for editing are generated.

\subsection{EMMET Derivation}\label{sec:derivation}
 Let $K_0 = [k^0_1 \hspace{4pt}| k^0_2 \hspace{4pt}| \dots |\hspace{4pt}k^0_N]$ represent $N$ key-vectors whose representations we want to preserve. Additionally, let $k^e_1, k^e_2 \dots k^e_E$ represent key-vectors for $E$ facts we want to edit in the model at the same time. Then according to the preservation-memorization objective, we want to find new weights $\hat{W}$ for a weight matrix $W_0$ such that:

\begin{equation}
\begin{aligned}
    \underset{\hat{W}}{\operatorname{argmin}} \hspace{4pt} \underbrace{\left\| \hat{W} K_0 - W_0 K_0 \right\|}_{\text{preservation}} \hspace{8pt}  \text{    s.t. } \\ \hspace{8pt} \underbrace{\hat{W} k^e_i = v^e_i \hspace{8pt} \forall i \in [1, 2 \dots E]}_{\text{memorization}}
\end{aligned}
\end{equation}

As can be seen in the above equation, the preservation of representations happens in the first term whereas memorization of all the new facts are forced using an equality constraint in the second term. The above equation is solved using lagrange-multipliers. The Lagrangian for the above equation for multiple equality constraints requires a summation of lagrange multipliers and equals:
\begin{equation}
\begin{aligned}
 L(\hat{W}, \lambda_i) = \frac{1}{2}\hat{W} K_0 K^T_0 \hat{W}^T - \hat{W} K_0 K^T_0 W^T_0  \\
 + \frac{1}{2}W_0 K_0 K^T_0 W^T_0 - \sum_{i=1}^E \lambda_i^T (\hat{W}k^e_i - v^e_i) 
\end{aligned}
\end{equation}

% As can be seen in the above equation, the preservation of representations happens in the first term whereas memorization of all the new facts are forced using an equality constrain in the second term. 
To solve the system of equations, we put $\frac{\delta L}{\delta \hat{W}} = 0 $ to get:
\begin{equation}
    \hat{W} K_0 K_0^T = W_0 K_0 K_0^T + \sum_{i=1}^E \lambda_i k^{e^T}_i\\
\end{equation}

which is same as:
\begin{equation}
 \hspace{8pt} (\hat{W} - W_0) K_0 K_0^T = \sum_{i=1}^E \lambda_i k^{e^T}_i = \Lambda K_E^T\\
\end{equation}

where $\Lambda = [\lambda_1 \hspace{4pt}| \lambda_2 \hspace{4pt}| \dots |\hspace{4pt}\lambda_E]$ and $K_E = [k^e_1 \hspace{4pt}| k^e_2 \hspace{4pt}| \dots |\hspace{4pt}k^e_E]$. Here, $\Lambda$ and $K_E$ are matrices created using a row of vectors. We set $K_0K_0^T = C_0$ (assuming that $C_0$ is invertible\footnote{In practice, we find that $C_0$ is always invertible as long as the number of key-vectors in $K_0$ are large enough}) to get the update equation of EMMET:
\begin{equation}\label{eq:emmet_intermediate}
 \hspace{8pt}\hat{W} = W_0 + \Lambda K_E^TC^{-1}_0\\
\end{equation}

where $\Lambda = [\lambda_1 \hspace{4pt}| \lambda_2 \hspace{4pt}| \dots |\hspace{4pt}\lambda_E]$, $K_E = [k^e_1 \hspace{4pt}| k^e_2 \hspace{4pt}| \dots |\hspace{4pt}k^e_E]$ and $C_0 = K_0K_0^T$. 

The unknown matrix of lagrange multipliers ($\Lambda$) can be found using the constraint $\hat{W}K_E = V_E$ in the previous equation. It comes out to be: 

\begin{equation}
    \Lambda = \left ( V_E - W_0 K_E \right ) \left ( K_E^T C^{-1}_0 K_E \right )^{-1}
\end{equation}

Replacing the above equation in equation \ref{eq:emmet_intermediate} gives us the update equation for EMMET:

\begin{equation}\label{eq:emmet_final}
\begin{aligned}
 &\hat{W} = W_0 + \Delta \hspace{10pt} \text{where} 
 \\ \Delta &= \left ( V_E - W_0 K_E \right ) \left ( K_E^T C^{-1}_0 K_E \right )^{-1} K_E^TC^{-1}_0 
 \end{aligned}
\end{equation}

\subsection{EMMET - MEMIT Hyperparameter Comparison}\label{sec:appendix-hparams}
\begin{figure*}
    \centering
    \begin{subfigure}{.32\textwidth}
        \centering
        \includegraphics[width=\linewidth]{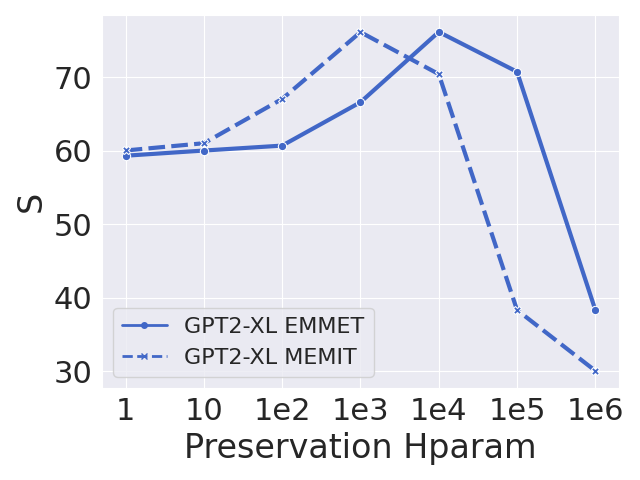}
        \caption{GPT2-XL}
    \end{subfigure}%
    \begin{subfigure}{.32\textwidth}
        \centering
        \includegraphics[width=\linewidth]{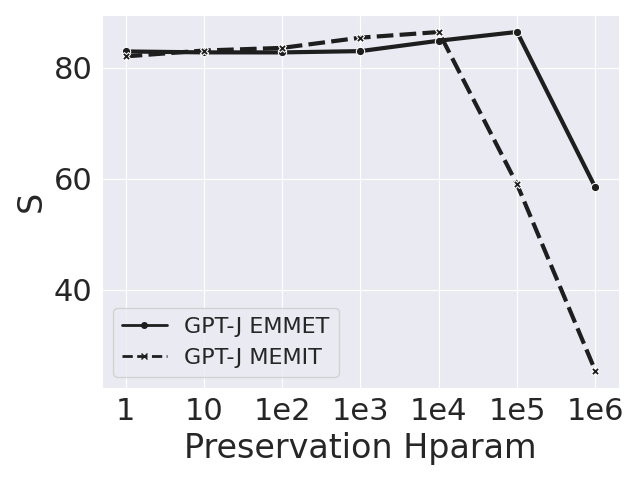}
        \caption{GPT-J}
    \end{subfigure}%
    \begin{subfigure}{.32\textwidth}
        \centering
        \includegraphics[width=\linewidth]{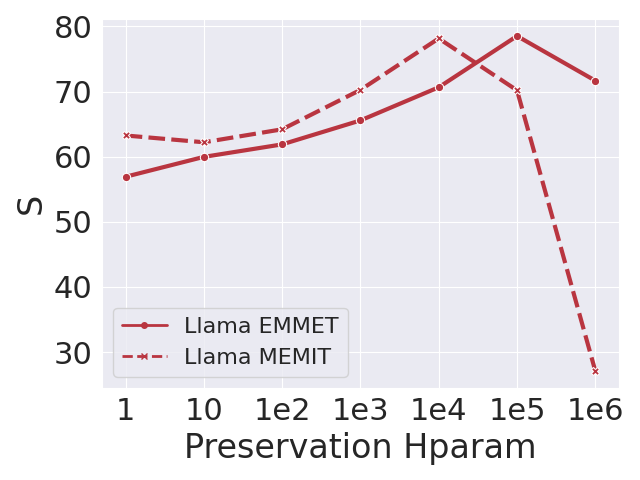}
        \caption{Llama-2-7b}
    \end{subfigure}
    
    \caption{Comparison between EMMET and MEMIT for different hyperparameter values for the metric of Score. Hyperparameter controls the weight of preservation term over memorization term.}
    \label{fig:hparams_S}
\end{figure*}

\begin{figure*}
    \centering
    \begin{subfigure}{.32\textwidth}
        \centering
        \includegraphics[width=\linewidth]{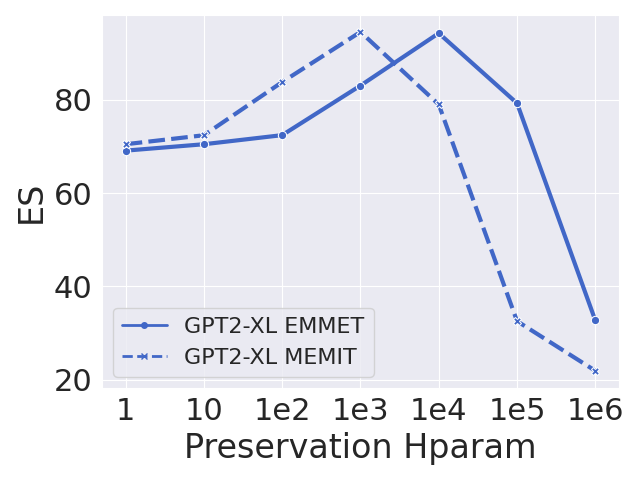}
        \caption{GPT2-XL}
    \end{subfigure}%
    \begin{subfigure}{.32\textwidth}
        \centering
        \includegraphics[width=\linewidth]{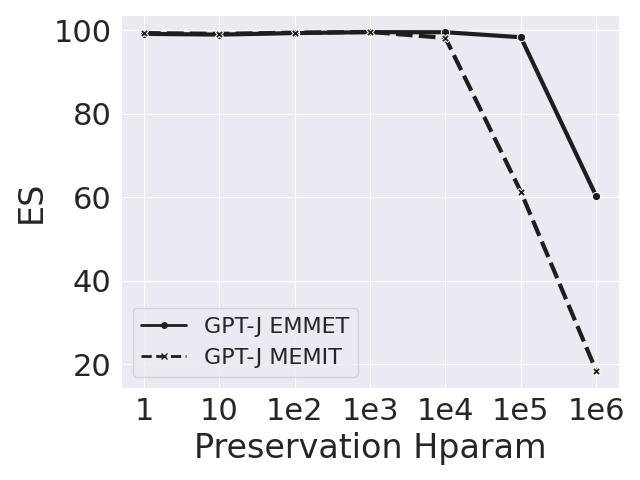}
        \caption{GPT-J}
    \end{subfigure}%
    \begin{subfigure}{.32\textwidth}
        \centering
        \includegraphics[width=\linewidth]{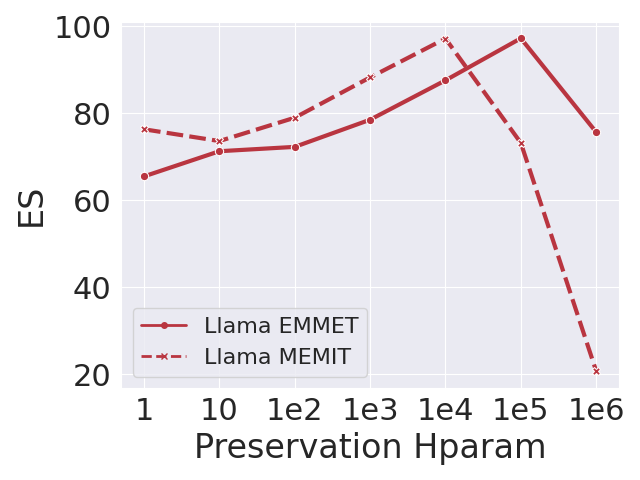}
        \caption{Llama-2-7b}
    \end{subfigure}
    
    \caption{Comparison between EMMET and MEMIT for different hyperparameter values for the metric of Efficacy Score.}
    \label{fig:hparams_ES}
\end{figure*}

\begin{figure*}
    \centering
    \begin{subfigure}{.32\textwidth}
        \centering
        \includegraphics[width=\linewidth]{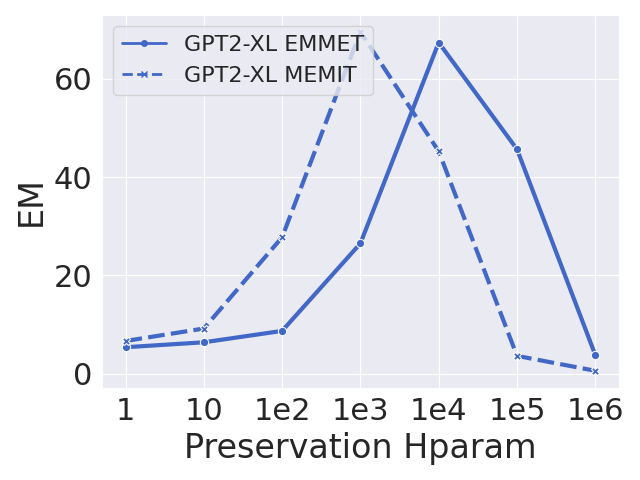}
        \caption{GPT2-XL}
    \end{subfigure}%
    \begin{subfigure}{.32\textwidth}
        \centering
        \includegraphics[width=\linewidth]{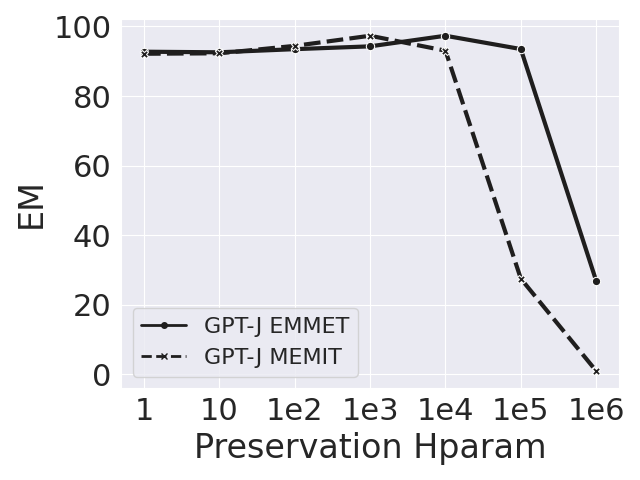}
        \caption{GPT-J}
    \end{subfigure}%
    \begin{subfigure}{.32\textwidth}
        \centering
        \includegraphics[width=\linewidth]{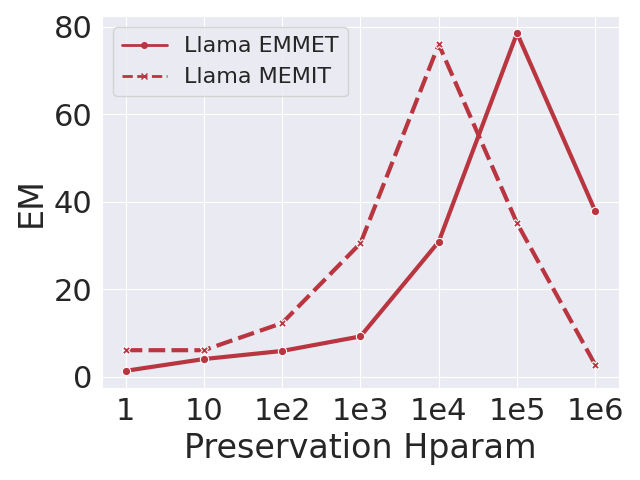}
        \caption{Llama-2-7b}
    \end{subfigure}
    
    \caption{Comparison between EMMET and MEMIT for different hyperparameter values for the metric of Efficacy Magnitude.}
    \label{fig:hparams_EM}
\end{figure*}

\begin{figure*}
    \centering
    \begin{subfigure}{.32\textwidth}
        \centering
        \includegraphics[width=\linewidth]{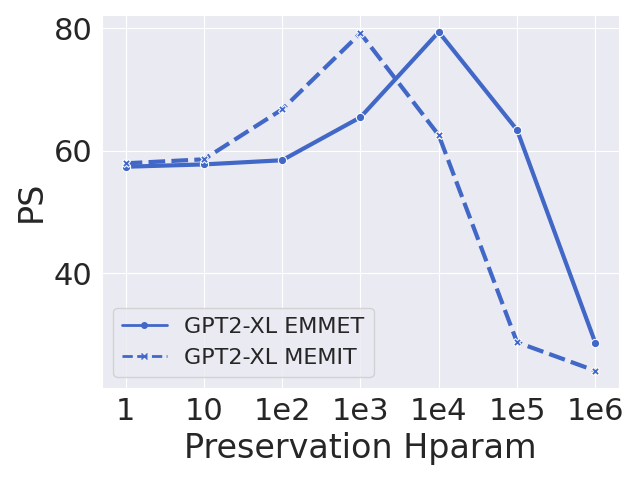}
        \caption{GPT2-XL}
    \end{subfigure}%
    \begin{subfigure}{.32\textwidth}
        \centering
        \includegraphics[width=\linewidth]{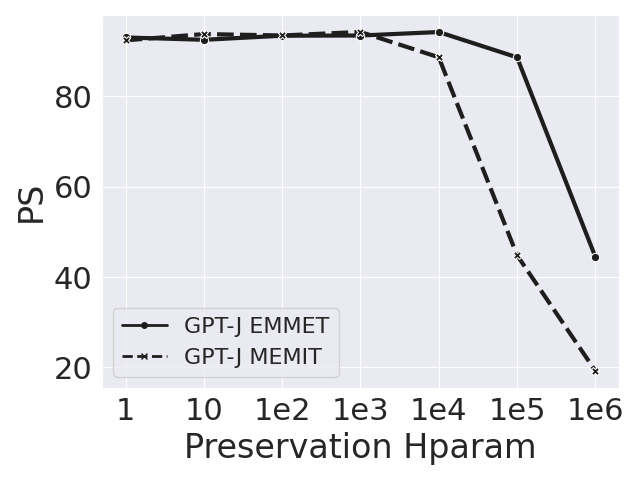}
        \caption{GPT-J}
    \end{subfigure}%
    \begin{subfigure}{.32\textwidth}
        \centering
        \includegraphics[width=\linewidth]{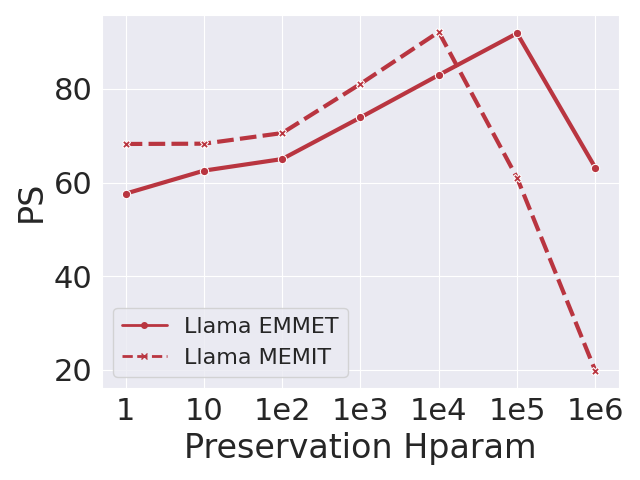}
        \caption{Llama-2-7b}
    \end{subfigure}
    
    \caption{Comparison between EMMET and MEMIT for different hyperparameter values for the metric of Paraphrase Score.}
    \label{fig:hparams_PS}
\end{figure*}

\begin{figure*}
    \centering
    \begin{subfigure}{.32\textwidth}
        \centering
        \includegraphics[width=\linewidth]{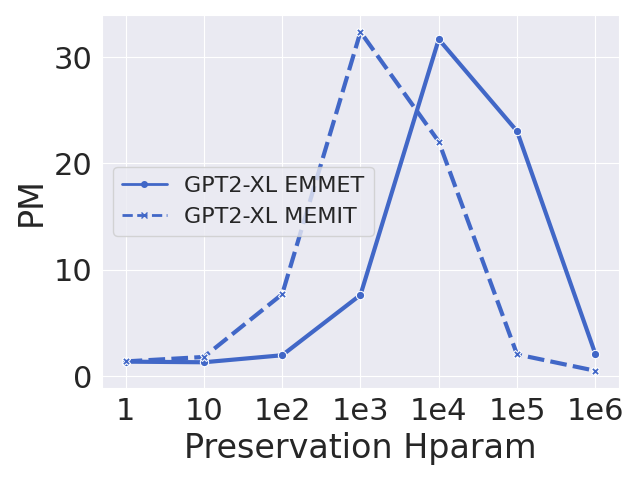}
        \caption{GPT2-XL}
    \end{subfigure}%
    \begin{subfigure}{.32\textwidth}
        \centering
        \includegraphics[width=\linewidth]{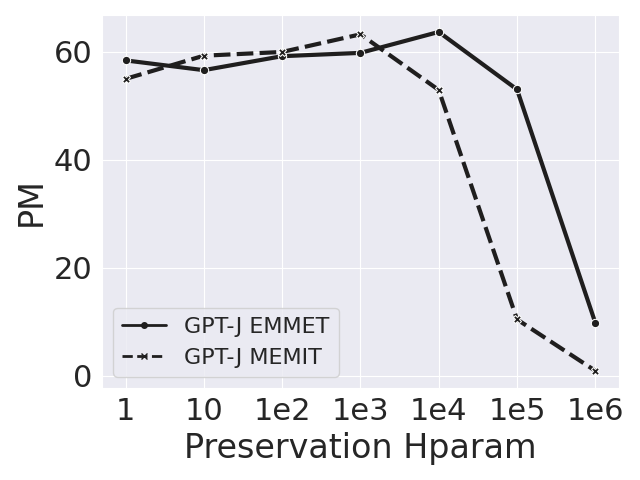}
        \caption{GPT-J}
    \end{subfigure}%
    \begin{subfigure}{.32\textwidth}
        \centering
        \includegraphics[width=\linewidth]{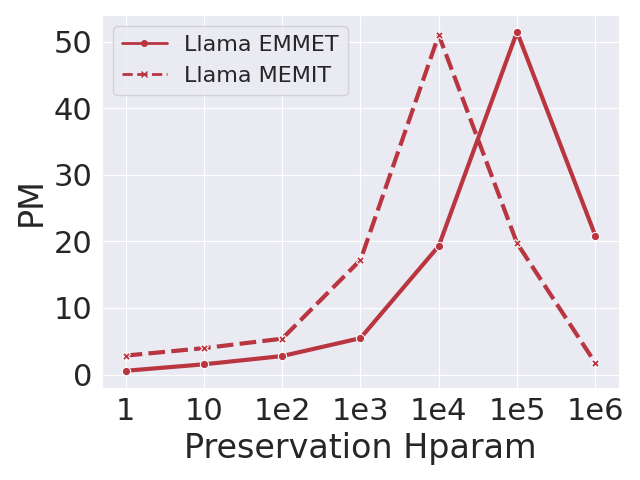}
        \caption{Llama-2-7b}
    \end{subfigure}
    
    \caption{Comparison between EMMET and MEMIT for different hyperparameter values for the metric of Paraphrase Magnitude.}
    \label{fig:hparams_PM}
\end{figure*}

\begin{figure*}
    \centering
    \begin{subfigure}{.32\textwidth}
        \centering
        \includegraphics[width=\linewidth]{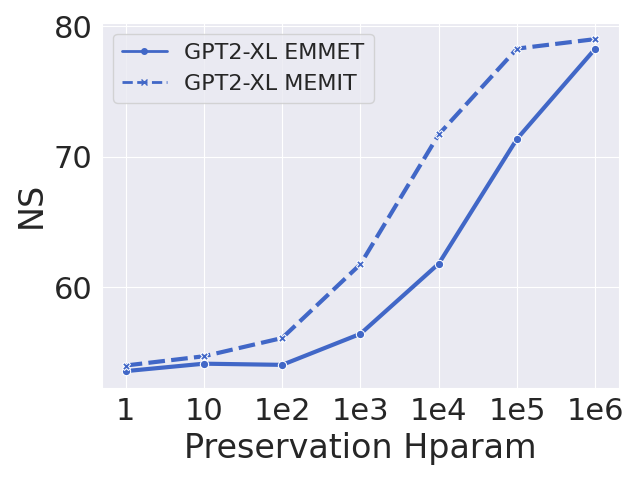}
        \caption{GPT2-XL}
    \end{subfigure}%
    \begin{subfigure}{.32\textwidth}
        \centering
        \includegraphics[width=\linewidth]{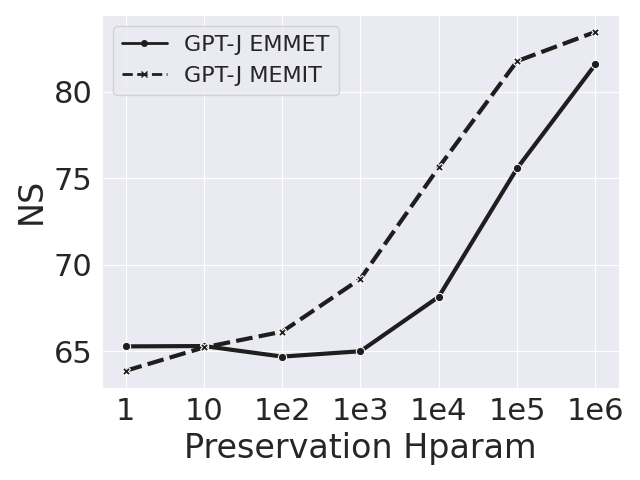}
        \caption{GPT-J}
    \end{subfigure}%
    \begin{subfigure}{.32\textwidth}
        \centering
        \includegraphics[width=\linewidth]{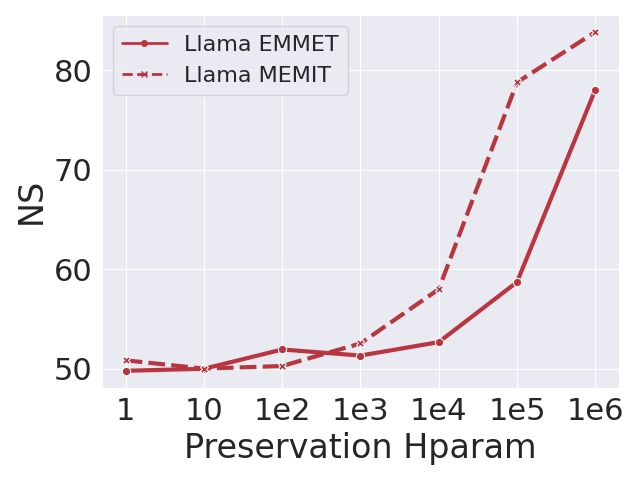}
        \caption{Llama-2-7b}
    \end{subfigure}
    
    \caption{Comparison between EMMET and MEMIT for different hyperparameter values for the metric of Neighborhood Score.}
    \label{fig:hparams_NS}
\end{figure*}

\begin{figure*}
    \centering
    \begin{subfigure}{.32\textwidth}
        \centering
        \includegraphics[width=\linewidth]{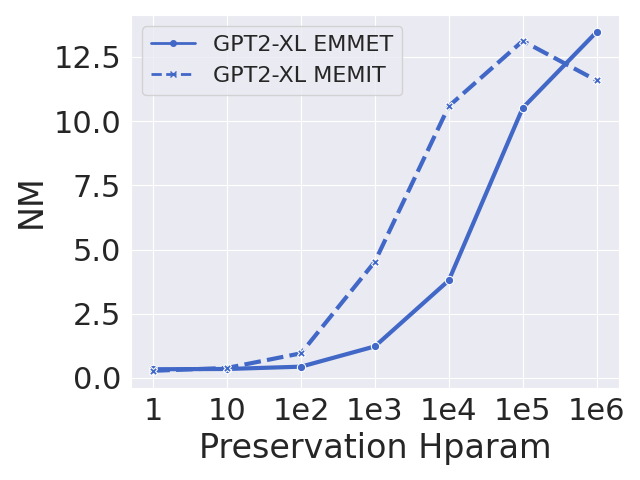}
        \caption{GPT2-XL}
    \end{subfigure}%
    \begin{subfigure}{.32\textwidth}
        \centering
        \includegraphics[width=\linewidth]{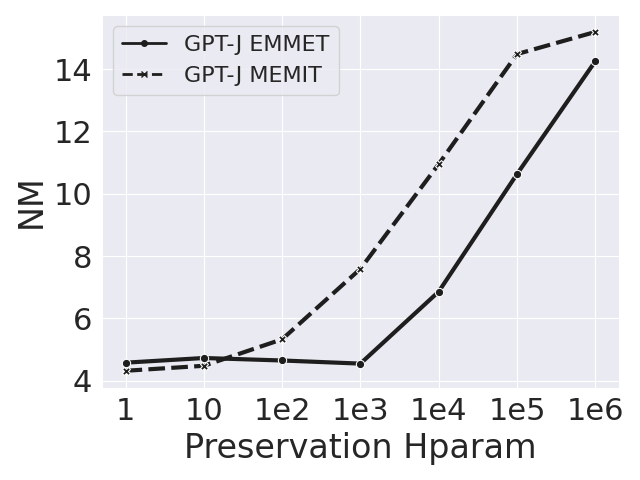}
        \caption{GPT-J}
    \end{subfigure}%
    \begin{subfigure}{.32\textwidth}
        \centering
        \includegraphics[width=\linewidth]{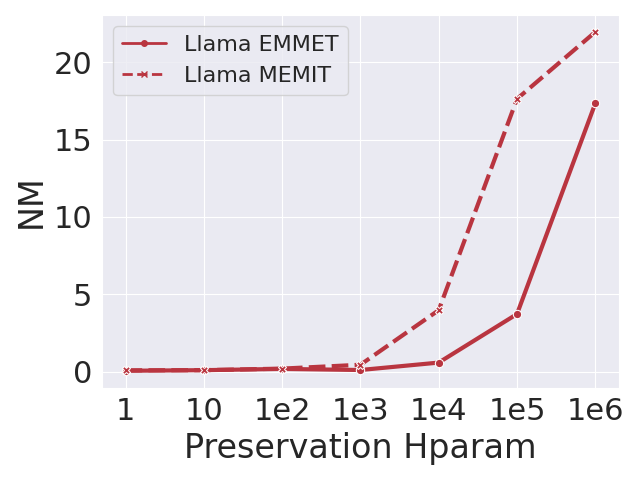}
        \caption{Llama-2-7b}
    \end{subfigure}
    
    \caption{Comparison between EMMET and MEMIT for different hyperparameter values for the metric of Neighborhood Magnitude.}
    \label{fig:hparams_NM}
\end{figure*}

\begin{figure*}
    \centering
    \begin{subfigure}{.32\textwidth}
        \centering
        \includegraphics[width=\linewidth]{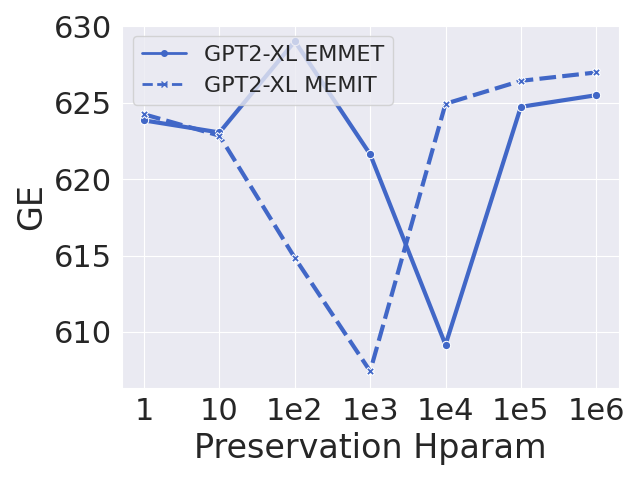}
        \caption{GPT2-XL}
    \end{subfigure}%
    \begin{subfigure}{.32\textwidth}
        \centering
        \includegraphics[width=\linewidth]{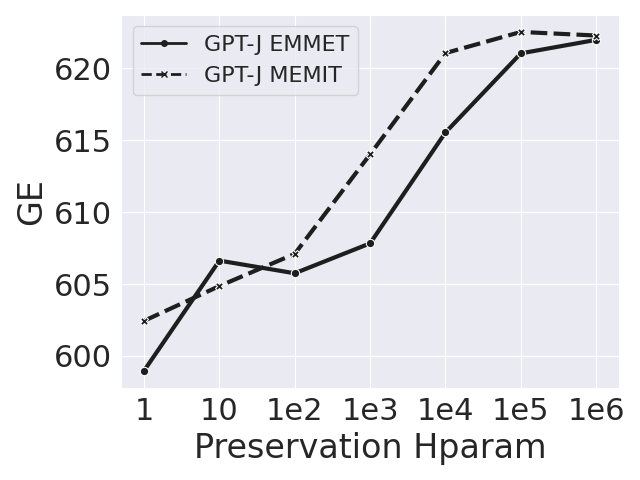}
        \caption{GPT-J}
    \end{subfigure}%
    \begin{subfigure}{.32\textwidth}
        \centering
        \includegraphics[width=\linewidth]{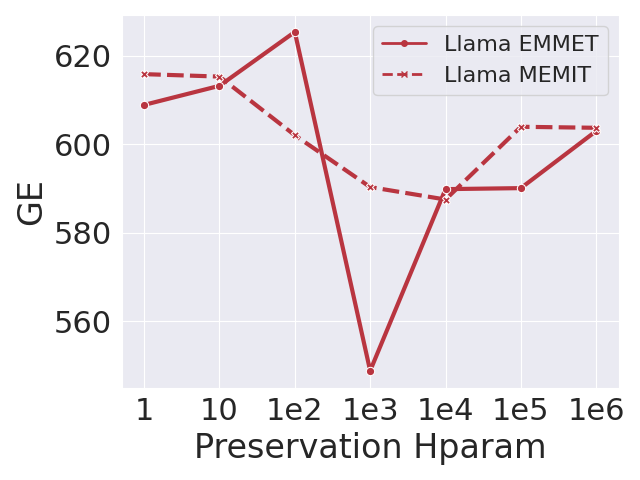}
        \caption{Llama-2-7b}
    \end{subfigure}
    
    \caption{Comparison between EMMET and MEMIT for different hyperparameter values for the metric of Generation Entropy.}
    \label{fig:hparams_GE}
\end{figure*}

Figures \ref{fig:hparams_S} - \ref{fig:hparams_GE} present the comparison between EMMET and MEMIT for different hyperparameter values. The hyperparameter corresponds to the preservation term in the preservation memorization objective (equation \ref{eq:memit_objective}). The figures show that both algorithm reach the same peak performance (Figure \ref{fig:hparams_S}) across all models, but at different hyperparameter values. MEMIT reaches peak performance at lower hyperparameter values, whereas EMMET needs a larger weight for preservation to reach similar performance. This makes sense as EMMET works with a much stronger memorization constraint and thus requires larger weight to preserve the model by the same amount. 

\newpage

\subsection{EMMET and MEMIT Downstream Performance Comparison}

\begin{figure}[htbp]
    \begin{subfigure}{.24\textwidth}
        \includegraphics[width=\linewidth]{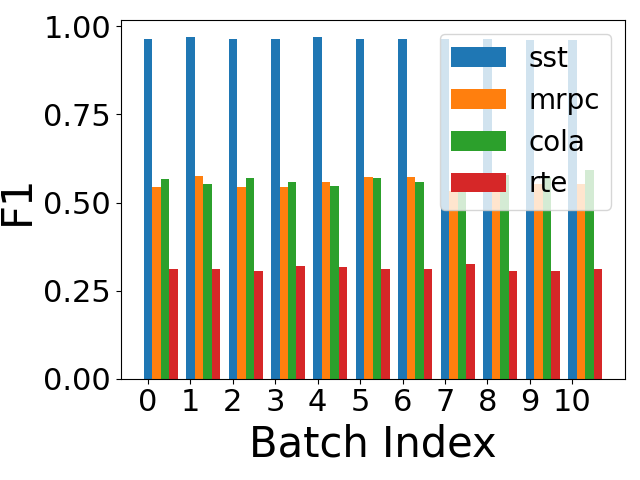}
        \caption{EMMET}
        \label{fig:memit_gptj:edit_score}
    \end{subfigure}%
    \begin{subfigure}{.24\textwidth}
        \includegraphics[width=\linewidth]{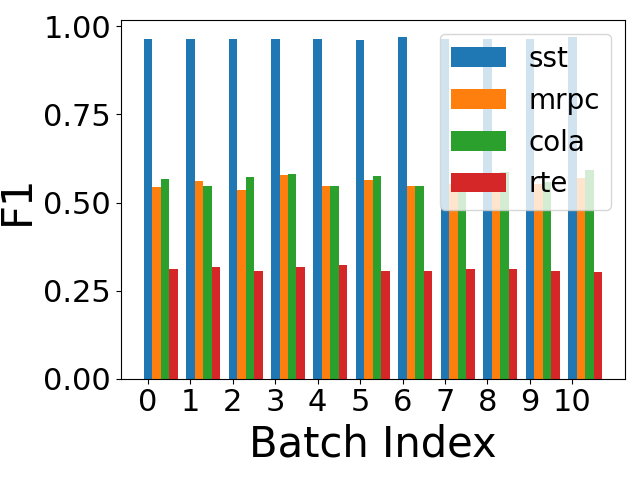}
        \caption{MEMIT}
        \label{fig:memit_gptj:downstream}
    \end{subfigure}
    \caption{Model - Llama2-7b. Batch size 4.}
    \label{fig:downstream_4}
\end{figure}

\begin{figure}[htbp]
    \begin{subfigure}{.24\textwidth}
        \includegraphics[width=\linewidth]{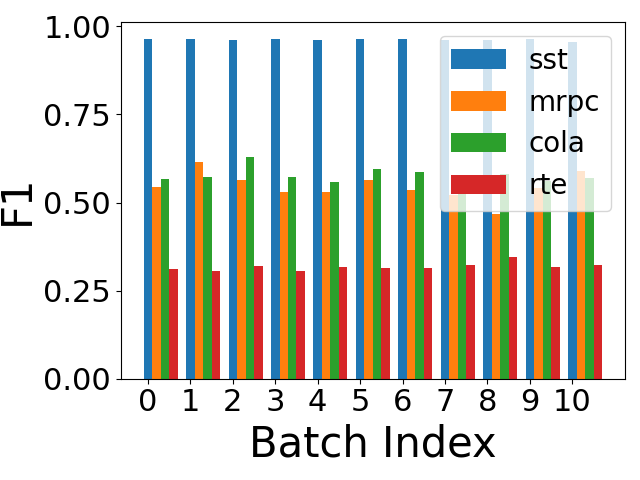}
        \caption{EMMET}
        \label{fig:memit_gptj:edit_score}
    \end{subfigure}%
    \begin{subfigure}{.24\textwidth}
        \includegraphics[width=\linewidth]{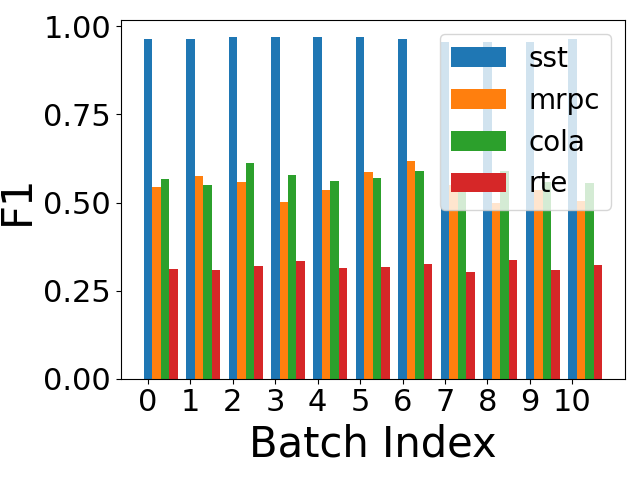}
        \caption{MEMIT}
        \label{fig:memit_gptj:downstream}
    \end{subfigure}
    \caption{Model - Llama2-7b. Batch size 16.}
    \label{fig:downstream_16}
\end{figure}

\begin{figure}[htbp]
    \begin{subfigure}{.24\textwidth}
        \includegraphics[width=\linewidth]{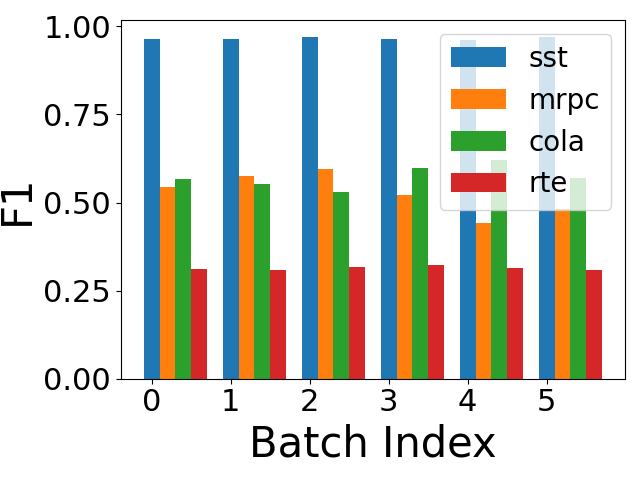}
        \caption{EMMET}
        \label{fig:memit_gptj:edit_score}
    \end{subfigure}%
    \begin{subfigure}{.24\textwidth}
   
        \includegraphics[width=\linewidth]{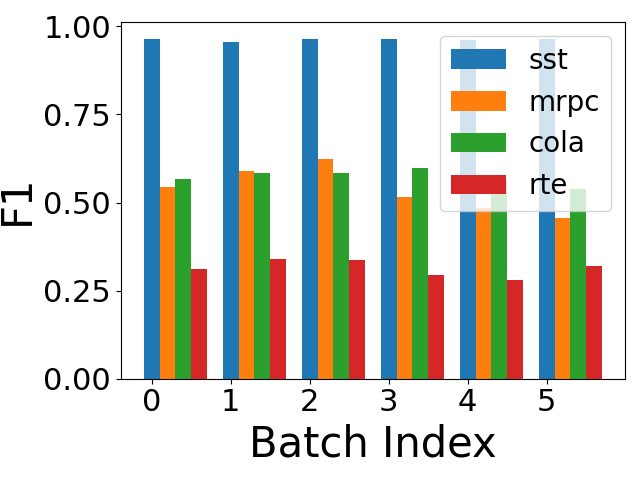}
        \caption{MEMIT}
        \label{fig:memit_gptj:downstream}
    \end{subfigure}
    \caption{Model - Llama2-7b. Batch size 64.}
    \label{fig:downstream_64}
\end{figure}

\begin{figure}
    \begin{subfigure}{.24\textwidth}
        \includegraphics[width=\linewidth]{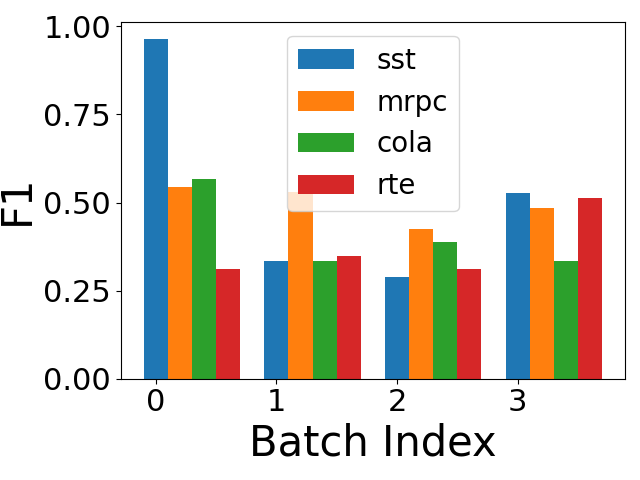}
        \caption{EMMET}
        \label{fig:memit_gptj:edit_score}
    \end{subfigure}%
    \begin{subfigure}{.24\textwidth}
        \includegraphics[width=\linewidth]{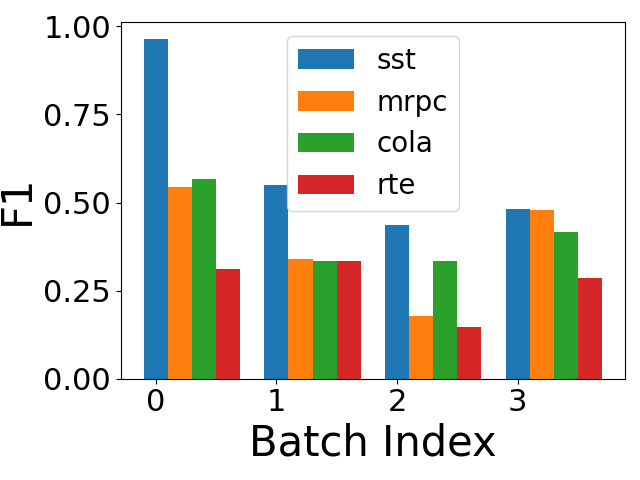}
        \caption{MEMIT}
        \label{fig:memit_gptj:downstream}
    \end{subfigure}
    \caption{Model - Llama2-7b. Batch size 1024.}
    \label{fig:downstream_1024}
\end{figure}

\begin{figure}
    \begin{subfigure}{.24\textwidth}
        \includegraphics[width=\linewidth]{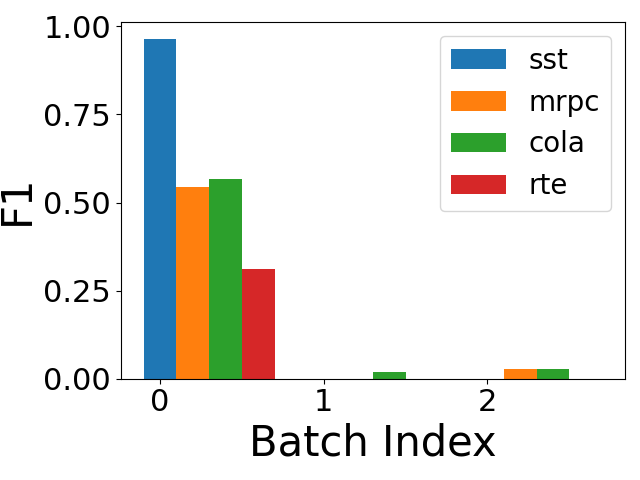}
        \caption{EMMET}
        \label{fig:memit_gptj:edit_score}
    \end{subfigure}%
    \begin{subfigure}{.24\textwidth}
        \includegraphics[width=\linewidth]{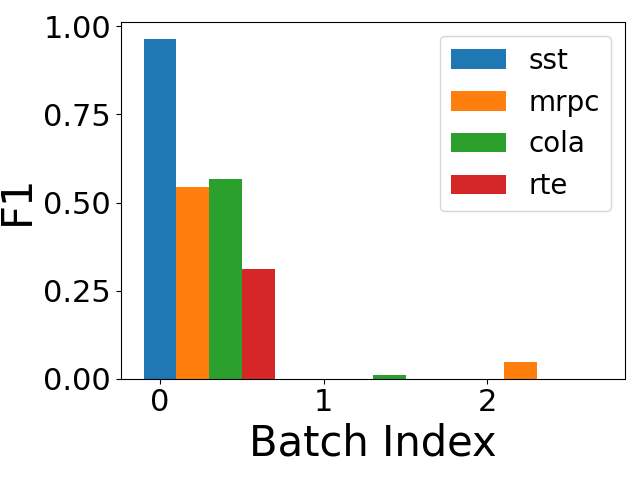}
        \caption{MEMIT}
        \label{fig:memit_gptj:downstream}
    \end{subfigure}
    \caption{Model - Llama2-7b. Batch size 4096.}
    \label{fig:downstream_4096}
\end{figure}

\begin{figure}
    \begin{subfigure}{.24\textwidth}
        \includegraphics[width=\linewidth]{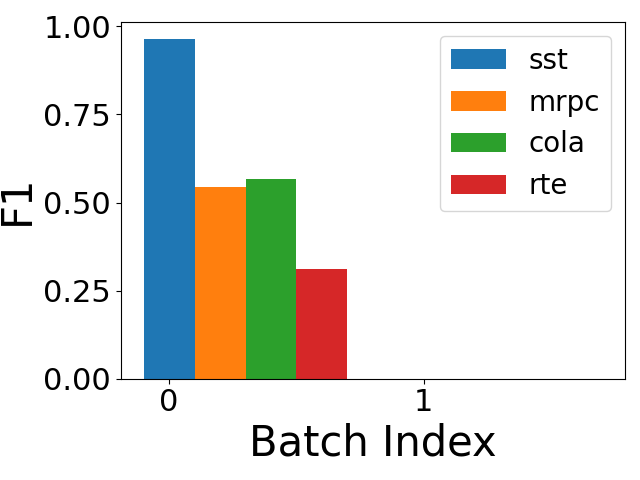}
        \caption{EMMET}
        \label{fig:memit_gptj:edit_score}
    \end{subfigure}%
    \begin{subfigure}{.24\textwidth}
        \includegraphics[width=\linewidth]{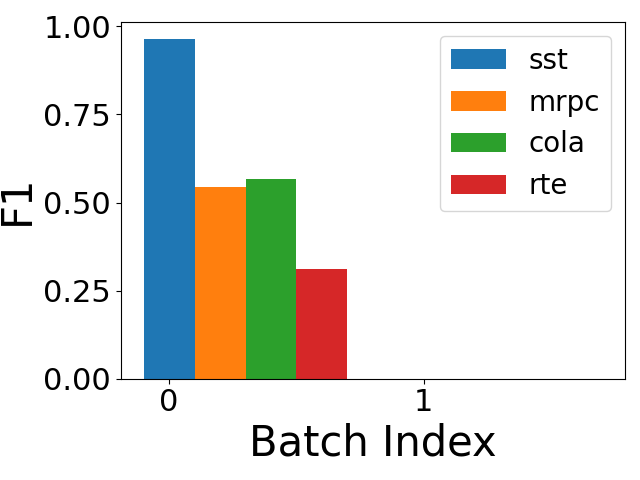}
        \caption{MEMIT}
        \label{fig:memit_gptj:downstream}
    \end{subfigure}
    \caption{Model - Llama2-7b. Batch size 10k.}
    \label{fig:downstream_10k}
\end{figure}

\end{document}